\PassOptionsToPackage{prologue,table}{xcolor}
\documentclass[sigconf, nonacm]{acmart}

\newcommand\vldbdoi{XX.XX/XXX.XX}
\newcommand\vldbpages{XXX-XXX}
\newcommand\vldbvolume{14}
\newcommand\vldbissue{1}
\newcommand\vldbyear{2020}
\newcommand\vldbauthors{\authors}
\newcommand\vldbtitle{\shorttitle} 
\newcommand\vldbavailabilityurl{URL_TO_YOUR_ARTIFACTS}
\newcommand\vldbpagestyle{plain} 
\usepackage{multirow}
\usepackage[normalem]{ulem}
\useunder{\uline}{\ul}{}
\usepackage{rotating}
\usepackage{subcaption}
\usepackage{enumitem}
\usepackage[ruled,linesnumbered]{algorithm2e}

\begin{document}
\begin{sloppypar}

\title{EffiCANet: Efficient Time Series Forecasting with Convolutional Attention}

\settopmatter{authorsperrow=4}
\author{Xinxing Zhou}
\affiliation{%
  \institution{College of CS} 
  \institution{Nankai University}
}
\email{zhouxinxing@dbis.nankai.edu.cn}

\author{Jiaqi Ye}
\affiliation{%
  \institution{College of CS} 
  \institution{Nankai University}
}
\email{yjq@mail.nankai.edu.cn}

\author{Shubao Zhao}
\affiliation{%
  \institution{Digital Research Institute}
  \institution{of ENN Group}
}
\email{zhaoshubao@enn.cn}

\author{Ming Jin}
\affiliation{%
    \institution{School of ICT} 
  \institution{Griffith University}
}
\email{mingjinedu@gmail.com}

\author{Chengyi Yang}
\affiliation{%
  \institution{Digital Research Institute}
  \institution{of ENN Group}
}
\email{yangchengyia@enn.cn}

\author{Yanlong Wen}
\authornote{Corresponding author}
\affiliation{%
  \institution{College of CS} 
  \institution{Nankai University}
}
\email{wenyl@nankai.edu.cn}

\author{Xiaojie Yuan}
\affiliation{%
  \institution{College of CS} 
  \institution{Nankai University}
}
\email{yuanxj@nankai.edu.cn}

\begin{abstract}
The exponential growth of multivariate time series data from sensor networks in domains like industrial monitoring and smart cities requires efficient and accurate forecasting models. Current deep learning methods often fail to adequately capture long-range dependencies and complex inter-variable relationships, especially under real-time processing constraints. These limitations arise as many models are optimized for either short-term forecasting with limited receptive fields or long-term accuracy at the cost of efficiency. Additionally, dynamic and intricate interactions between variables in real-world data further complicate modeling efforts. To address these limitations, we propose EffiCANet, an Efficient Convolutional Attention Network designed to enhance forecasting accuracy while maintaining computational efficiency. EffiCANet integrates three key components: (1) a Temporal Large-kernel Decomposed Convolution (TLDC) module that captures long-term temporal dependencies while reducing computational overhead; (2) an Inter-Variable Group Convolution (IVGC) module that captures complex and evolving relationships among variables; and (3) a Global Temporal-Variable Attention (GTVA) mechanism that prioritizes critical temporal and inter-variable features. Extensive evaluations across nine benchmark datasets show that EffiCANet achieves the maximum reduction of 10.02\% in MAE over state-of-the-art models, while cutting computational costs by 26.2\% relative to conventional large-kernel convolution methods, thanks to its efficient decomposition strategy.
\end{abstract}

\maketitle

\pagestyle{\vldbpagestyle}
\begingroup\small\noindent\raggedright\textbf{PVLDB Reference Format:}\\
\vldbauthors. \vldbtitle. PVLDB, \vldbvolume(\vldbissue): \vldbpages, \vldbyear.\\
\href{https://doi.org/\vldbdoi}{doi:\vldbdoi}
\endgroup
\begingroup
\renewcommand\thefootnote{}\footnote{\noindent
This work is licensed under the Creative Commons BY-NC-ND 4.0 International License. Visit \url{https://creativecommons.org/licenses/by-nc-nd/4.0/} to view a copy of this license. For any use beyond those covered by this license, obtain permission by emailing \href{mailto:info@vldb.org}{info@vldb.org}. Copyright is held by the owner/author(s). Publication rights licensed to the VLDB Endowment. \\
\raggedright Proceedings of the VLDB Endowment, Vol. \vldbvolume, No. \vldbissue\ %
ISSN 2150-8097. \\
\href{https://doi.org/\vldbdoi}{doi:\vldbdoi} \\
}\addtocounter{footnote}{-1}\endgroup

\ifdefempty{\vldbavailabilityurl}{}{
\vspace{.3cm}
\begingroup\small\noindent\raggedright\textbf{PVLDB Artifact Availability:}\\
The source code, data, and/or other artifacts have been made available at \url{\vldbavailabilityurl}.
\endgroup
}

\section{Introduction}\label{Introduction}
As we advance into an era dominated by AI-driven systems and ubiquitous Internet of Things (IoT), time series data has become a cornerstone of modern industries ~\cite{cheng2023weakly, zhao2023multiple, sylligardos2023choose}. From intelligent manufacturing and smart cities to precision medicine, vast sensor networks continuously generate complex multivariate data streams, driving real-time insights for proactive monitoring, resource scheduling, and predictive maintenance ~\cite{farahani2025time, zhang2024towards, qin2023federated, wu2023interpretable}. This data surge underscores the need for advanced time series forecasting methods to enable data-driven decision-making and enhance operational intelligence across diverse emerging applications. 

Traditional time series forecasting techniques, including Transformer-based methods ~\cite{vaswani2017attention, nietime, zhou2021informer, zhou2022fedformer}, have demonstrated strong performance across various domains. However, while they excel at capturing long-term dependencies, they often struggle to balance computational efficiency with real-time processing in resource-constrained environments. In contrast, temporal convolutional networks (TCNs) ~\cite{lea2017temporal} are more efficient but suffer from limited receptive fields, restricting their ability to capture long-term patterns. Linear models ~\cite{zeng2023transformers, li2023revisiting}, on the other hand, offer simplicity and computational efficiency but lack the capacity to capture complex, non-linear dependencies. These trade-offs reveal a crucial gap in current research: there is a critical need for models that can balance efficiency with accurate forecasting over short and long horizons, especially for real-time decision-making. Bridging this gap is crucial for transforming the potential of IoT and AI-driven systems into actionable, timely, and reliable insights that optimize operations, enhance resource allocation, and foster innovation in sectors such as smart cities, precision manufacturing, and healthcare ~\cite{ding2023forecasting, piccialli2021artificial,walther2021systematic}.

\begin{figure}
  \centering
  \includegraphics[width=0.85\linewidth]{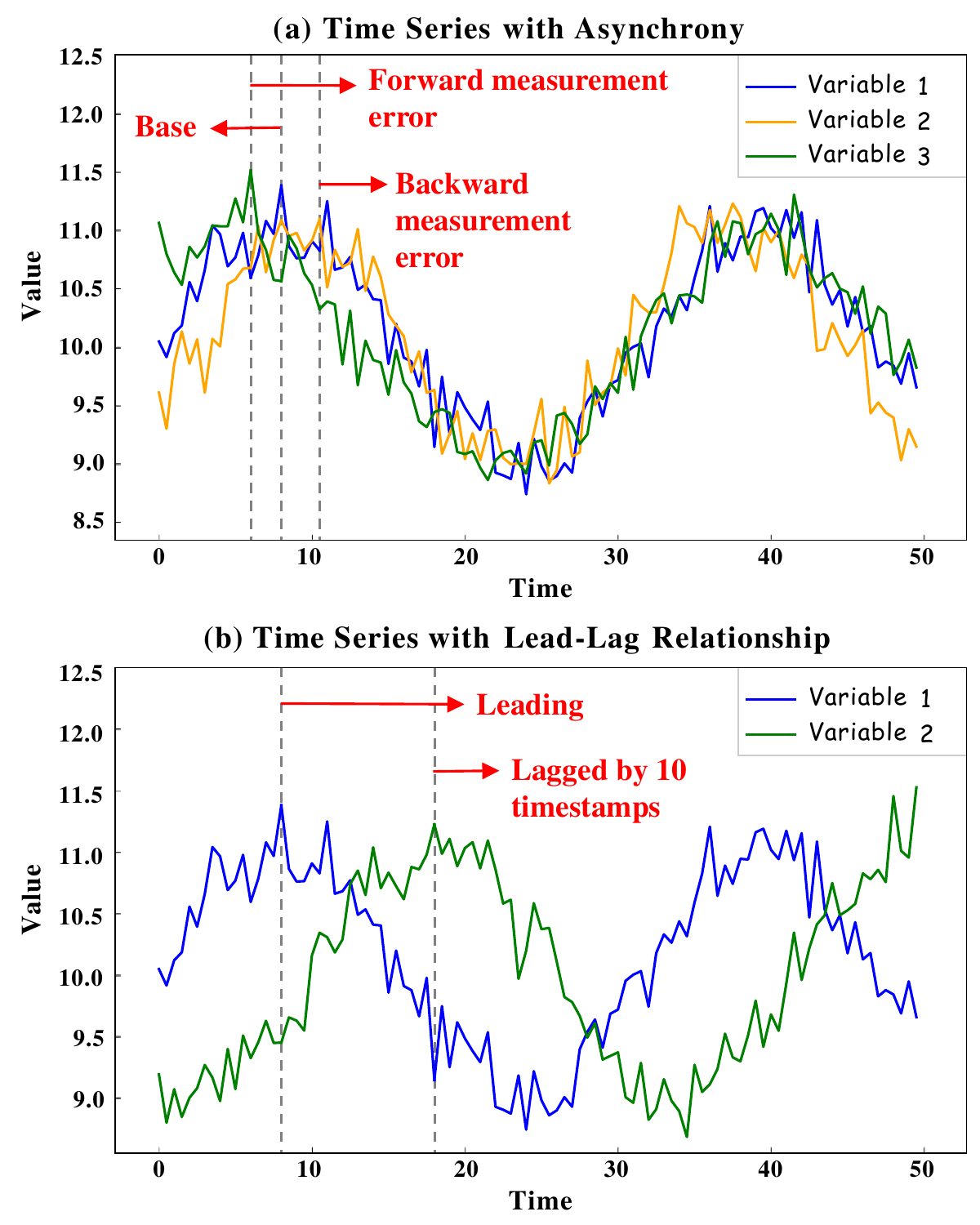}
  \caption{Illustration of asynchrony and lead-lag relationship in multiple time series: (a) Three variables exhibiting asynchronous behavior due to measurement errors, marked by vertical dashed lines. (b) A lead-lag relationship between two variables, with one leading the other by 10 timestamps.}
  \label{fig:illustration2}
  \vspace{-15pt}
\end{figure}

In real-world forecasting, time series data rarely exhibit independence. Interactions among variables are often complicated by asynchrony and lag effects. For example, \autoref{fig:illustration2}(a) shows three variables with asynchronous behavior due to measurement discrepancies, manifested as slight random shifts at specific timestamps. A common example can be found in climate monitoring, where temperature and humidity readings from different sensors may not align due to sensor calibration errors and environmental conditions, leading to misunderstandings regarding underlying environmental interactions. Besides, \autoref{fig:illustration2}(b) illustrates a lead-lag relationship, where one variable is influenced by the historical values of another. This scenario is often observed in supply chain management, where inventory levels may respond to changes in demand with a noticeable delay. Ignoring these relationships can cause models to miss important latent connections, reducing their ability to capture the dynamics of complex systems. Addressing these challenges is crucial for developing robust and reliable forecasting models that accurately reflect real-world behaviors.

In this paper, we study the problem of multivariate time series forecasting, focusing on efficiently capturing both short- and long-term dependencies, as well as the dynamic interactions between variables. However, existing approaches ~\cite{wu2021autoformer,nietime,zhao2024rethinking, zhao2024himtm} often struggle to balance accuracy, computational efficiency, and the complexity of inter-variable dynamics. Specifically, we identify two primary challenges in current methods that hinder their effectiveness in complex, real-time forecasting tasks.

\textbf{Challenge 1:} Efficiently capturing long-term dependencies. While mainstream architectures like Transformers are highly effective at learning long-term dependencies, their high computational cost limits their feasibility in real-time or resource-constrained scenarios. In contrast, convolutional models, such as TCNs, offer computationally efficient local pattern recognition but suffer from limited receptive fields, making long-term dependency capture challenging. Expanding the convolutional receptive field through large kernels ~\cite{luo2024moderntcn} partially alleviates this issue but increases computational complexity. The challenge lies in expanding the receptive field efficiently, enabling both short- and long-term forecasting without compromising speed or scalability.

\textbf{Challenge 2:} Modeling dynamic inter-variable dependencies with asynchronous and lagged relationships. In multivariate time series forecasting, variable dependencies are often complex and evolve dynamically. These relationships are further complicated by issues like sensor noise, time lags, and the heterogeneous nature of the variables. Existing models frequently emphasize either temporal dependencies or inter-variable relationships in isolation, failing to account for asynchronous or lagged interactions. This limitation leads to an incomplete understanding of variable interdependencies, especially when variables influence one another with time offsets. Developing methods that dynamically model these evolving relationships is crucial for accurate and robust forecasting. 

To address these challenges, we propose EffiCANet, an Efficient Convolutional Attention Network designed to capture both short- and long-term dependencies, while dynamically modeling complex inter-variable relationships. 

\textbf{Addressing Challenge 1:} To overcome the limitations of convolutional models in capturing long-term dependencies efficiently, we introduce the Temporal Large-kernel Decomposed Convolution (TLDC) module. This module innovatively decomposes large kernels into a series of smaller, computationally efficient convolutions, significantly expanding the receptive field without a proportional increase in computational cost. Combined with an efficiently designed global temporal attention mechanism in the Global Temporal-Variable Attention (GTVA) module that adaptively highlights relevant time steps, our model proficiently captures both short- and long-term dependencies. This hybrid approach of convolution and attention effectively mitigates the computational bottlenecks of traditional large-kernel methods, making the model highly suitable for scenarios requiring low-latency predictions.

\textbf{Addressing Challenge 2:} To tackle the complexities of modeling dynamic inter-variable relationships, we present the Inter-Variable Group Convolution (IVGC) module. This module leverages group convolutions ~\cite{ma2018shufflenet} to capture inter-variable interactions over adjacent time windows, organizing variable-time window pairs into shared kernel groups for efficient processing. By sharing kernels across these groups, the model adapts to evolving dependencies. Additionally, a global variable attention mechanism in GTVA module assigns adaptive weights to each variable, dynamically responding to shifts in variable importance. This combined approach not only enhances the model's ability to learn intricate relationships in multivariate time series data but also ensures efficient computation, making it well-suited for real-time applications.

In summary, the key contributions of this paper are as follows:
\begin{itemize}
\item We identify the challenges of capturing long-term temporal dependencies and complex inter-variable relationships in resource-constrained environments and propose EffiCANet, a novel convolutional attention network designed to address these issues.

\item EffiCANet incorporates a large-kernel decomposed convolution module for efficient long-term temporal dependency modeling, a group convolution module for dynamic inter-variable dependency capture, and a global attention mechanism to enhance temporal and variable feature extraction.

\item Extensive experiments on nine public datasets demonstrate that EffiCANet consistently outperforms state-of-the-art methods in both forecasting accuracy and computational efficiency.
\end{itemize}

The remainder of this paper is organized as follows: Section \ref{Related Work} reviews the relevant literature. Section \ref{Methodology} details our proposed method. Section \ref{Experiments} describes the experimental setup and results. Finally, Section \ref{Conclusion} concludes the paper.

\section{Related Work}\label{Related Work}
\subsection{Convolution-based time series forecasting}
Convolutional neural networks (CNNs) have long been utilized in time series forecasting due to their ability to capture local temporal patterns efficiently. Despite the rise of Transformer-based models, CNNs remain relevant for their computational efficiency and adaptability ~\cite{livieris2020cnn,mehtab2022analysis,widiputra2021multivariate,wibawa2022time,hewage2020temporal,wan2019multivariate}. 

Convolutional techniques have evolved substantially over time. For instance, MLCNN ~\cite{cheng2020towards} utilizes convolutional layers to generate multi-level representations for near and distant future predictions. SCINet ~\cite{liu2022scinet} employs a recursive downsample-convolve-interact architecture, using multiple convolutional filters to extract temporal features from downsampled sequences. MICN ~\cite{wang2023micn} integrates multi-scale down-sampled and isometric convolutions to capture both local and global temporal patterns. TimesNet ~\cite{wutimesnet} transforms 1D time series into 2D tensors and applies inception blocks with 2D convolutional kernels to model period variations. Recently, the large-kernel convolution ~\cite{ding2022scaling} and separable convolution ~\cite{howard2017mobilenets}, originating from computer vision, have proven effective for time series analysis. ModernTCN ~\cite{luo2024moderntcn} extends the receptive field of traditional TCN architectures to enhance long-term dependency modeling, while ConvTimeNet ~\cite{cheng2024convtimenet} leverages depthwise and pointwise convolutions to model both global sequence and cross-variable dependencies.

While large kernels or deep architectures improve long-term dependency capture, they often lead to excessive parameter growth, computational overhead, and increased memory consumption. EffiCANet addresses these issues by using large-kernel decomposed convolutions, efficiently modeling long-term dependencies without sacrificing computational efficiency. 

\subsection{Attention-based time series forecasting}

Attention mechanisms, originally developed for tasks such as machine translation, have gained widespread adoption in time series forecasting. Their strength lies in dynamically focusing on relevant temporal patterns, enabling the modeling of both short-term and long-term dependencies.
\vspace{-3pt}
\subsubsection{Transformer Models} Transformer-based architectures ~\cite{vaswani2017attention} have become foundational in attention-based time series forecasting due to their proficiency in capturing long-range dependencies. Early Transformer models struggled with scalability in time series applications, prompting the development of specialized variants ~\cite{shiscaling}. LogTrans ~\cite{li2019enhancing} introduces sparse attention to reduce computational overhead, while Informer ~\cite{zhou2021informer} incorporates ProbSparse attention for enhanced scalability. Autoformer ~\cite{wu2021autoformer} and FEDformer ~\cite{zhou2022fedformer} adopt decomposition-based methods to separate trend and seasonal components. Pyraformer ~\cite{liu2021pyraformer} leverages hierarchical structures to balance efficiency and accuracy. Crossformer ~\cite{zhang2023crossformer} and PatchTST ~\cite{nietime} employs cross-dimensional attention and patch-based approaches to improve multivariate time series modeling. Recently, iTransformer ~\cite{liuitransformer} inverts the standard Transformer architecture to enhance multivariate correlation modeling by attending to variate tokens rather than temporal ones. Pathformer ~\cite{chen2023pathformer} adopts multi-scale modeling with adaptive pathways to capture temporal dynamics at various resolutions, while SAMformer ~\cite{ilbertsamformer} addresses overfitting issues through sharpness-aware optimization. TSLANet ~\cite{eldeletslanet} further refines attention mechanisms by incorporating adaptive spectral and interactive convolution blocks to better manage noise and multi-scale dependencies.

\subsubsection{Other Attention-Based Methods} Beyond the Transformer paradigm, other attention-based models have also shown promise in time series forecasting. These models combine attention mechanisms with various techniques to enhance the extraction of temporal patterns. MH-TAL ~\cite{fan2019multi} integrates temporal attention within RNNs to discover hidden patterns. RGDAN ~\cite{fan2024rgdan} incorporates a graph diffusion attention module for learning spatial dependencies and a temporal attention module to capture time-related correlations. FMamba ~\cite{ma2024fmamba} merges the selective state space model of Mamba ~\cite{gu2023mamba} with fast-attention techniques, optimizing temporal feature extraction while maintaining linear computational complexity.

Despite their success, attention-based models face challenges such as high computational cost and reliance on positional encodings, which may not fully capture complex temporal dynamics. EffiCANet addresses these limitations by integrating a more efficient attention mechanism that dynamically captures both temporal and inter-variable dependencies.
\vspace{-3pt}
\subsection{Linear models for time series forecasting}

Recently, linear models have re-emerged as powerful tools for time series forecasting due to their simplicity, interpretability, and effectiveness in capturing linear dependencies. Models such as DLinear and NLinear \cite{zeng2023transformers} first show that straightforward linear approaches can rival and occasionally surpass complex Transformer architectures in long-term forecasting. ~\cite{li2023revisiting} explore the impact of reversible normalization (RevIN) ~\cite{kim2021reversible} and channel independence (CI) on linear models, and propose RLinear, which consists of RevIN and a single linear layer that maps the input to the output time series. FITS \cite{xufits} applies a real discrete Fourier transform (RFT) and optional high-frequency filtering in the frequency domain.

However, linear models are inherently limited in their ability to capture the complex non-linear dependencies present in multivariate datasets. While they perform well on low-dimensional datasets, their performance deteriorates significantly on datasets with a larger number of variables, highlighting the need for more well-designed models like EffiCANet that can effectively capture both linear and non-linear dependencies.

\section{Methodology}\label{Methodology}

\begin{figure*}[h!]
  \centering
  \includegraphics[width=0.95\textwidth]{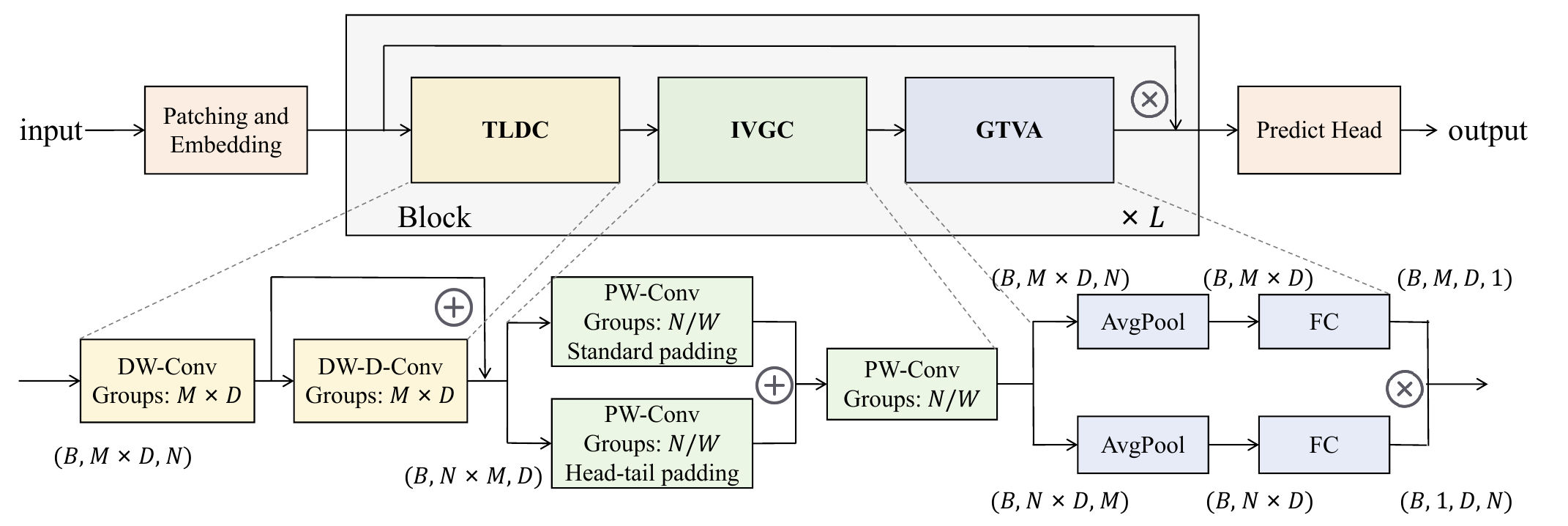}
  \caption{Overview of the EffiCANet architecture. The input is first processed through a patching and embedding layer, transforming raw data into a suitable feature space. The core structure consists of $L$ stacked Blocks, each containing three key modules: TLDC, IVGC, and GTVA. Within each Block, the output is iteratively refined by element-wise multiplication with its respective input, enhancing feature representations. Finally, the output is passed through a prediction head to generate the forecast.}
  \label{fig:overview}
  \vspace{-10pt}
\end{figure*}

\subsection{Problem Definition}
We consider a multivariate time series $\mathbf{X} \in \mathbb{R}^{M \times T}$, where $M$ represents the number of variables and $T$ denotes the length of the time series. Each element $\mathbf{X}_m^t \in \mathbb{R}$ corresponds to the value of the $m$-th variable at time step $t$. The sequence $\mathbf{X}_m \in \mathbb{R}^T$ represents the full time series for the $m$-th variable. Given a time interval $\Delta t$, the subsequence $\mathbf{X}^{t_0 : t_0+\Delta t} \in \mathbb{R}^{M \times \Delta t}$ refers to the segment of all variables between time steps $t_0$ and $t_0 + \Delta t$.

Our goal is to define a predictive model $\mathcal{F}$ that, given a historical window of the time series, can forecast the values for future time steps. Specifically, the model $\mathcal{F}$ takes a historical time window $\mathbf{X}^{t_0 - H : t_0}$, where $H$ is the window length, and predicts the next $\tau$ time steps, denoted as $\hat{\mathbf{X}}^{t_0 : t_0 + \tau}$. The relationship can be formalized as:
\vspace{-1pt}
\begin{equation}
    \hat{\mathbf{X}}^{t_0:t_0+\tau}=\mathcal{F}_\Phi (\mathbf{X}^{t_0-H:t_0})
\end{equation}

where $\Phi$ denotes the parameters of the model $\mathcal{F}$.

\subsection{Model Overview}

EffiCANet is designed to efficiently capture both temporal dependencies and inter-variable dynamics in multivariate time series forecasting. As shown in \autoref{fig:overview}, the model processes input data through a sequence of stacked blocks, each integrating large-kernel convolutions, group convolutions, and global attention mechanisms to capture complex relationships across both temporal and variable dimensions.

Given the multivariate time series input $\mathbf{X}_{\text{in}} = \mathbf{X}^{t_0-H:t_0} \in \mathbb{R}^{M \times H}$, we adopt a patching and embedding strategy inspired by ~\cite{luo2024moderntcn}. Initially, the input is reshaped via an unsqueeze operation into $\mathbf{X}_{\text{in}} \in \mathbb{R}^{M \times 1 \times H}$, introducing a channel dimension that enables independent processing of each variable. A convolutional stem layer then partitions the sequence into patches using a kernel size of $P$ and a stride of $S$, producing  $N=\lfloor\frac{H-P}{S}\rfloor+2$ overlapping patches.
This convolution maps the single input channel to $D$ output channels, resulting in an embedded tensor $\mathbf{X}_{\text{emb}} \in \mathbb{R}^{M \times D \times N}$, where $N$ represents the number of patches.

The model extracts features through a series of $L$ stacked blocks, each designed to iteratively refine the feature representations by capturing both localized and global temporal-variable patterns. Formally, the output of the $l$-th block, denoted as $\mathbf{Z}^{(l)}$, is computed as:
\vspace{-1pt}
\begin{equation}
    \mathbf{Z}^{(l)}=f_{\mathrm{Block}}^{(l)}(\mathbf{Z}^{(l-1)})
\end{equation}

where $\mathbf{Z}^{(0)} = \mathbf{X}_{\text{emb}}$, and $f_{\text{Block}}^{(l)}(\cdot)$ denotes the operations within the $l$-th block.

After processing through all $L$ blocks, the refined feature representation is passed to generate the final forecast $\hat{\mathbf{X}}^{t_0:t_0
+\tau} \in \mathbb{R}^{M \times \tau}$, predicting the values for the future time steps.

\subsection{Temporal Large-kernel Decomposed Convolution}

\begin{figure}
  \centering
  \includegraphics[width=\linewidth]{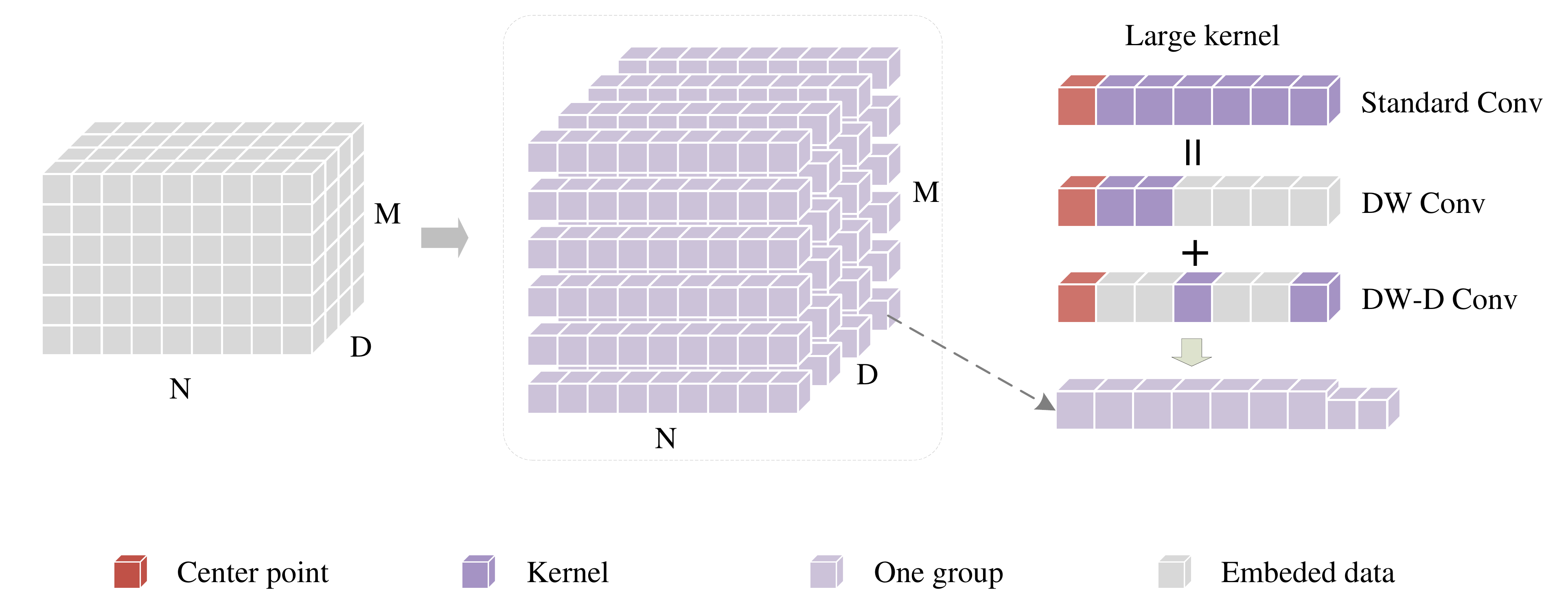}
  \caption{Illustration of the TLDC module. The input tensor, shaped as $(M, D, N)$, is processed through group convolution, segregating it into $M \times D$ groups corresponding to each variable-channel pair. Each group undergoes a decomposition of standard convolution into two stages: a depth-wise convolution (DW Conv) and a depth-wise dilated convolution (DW-D Conv). In this example, a DW Conv with a kernel size of 5 is followed by a DW-D Conv with the same kernel size and a dilation rate of 3, together simulating the effect of a large kernel size of 13. The combined outputs capture both short-term and long-term temporal dependencies effectively.}
  \label{fig:TLDC}
  \vspace{-10pt}
\end{figure}

The Temporal Large-kernel Decomposed Convolution (TLDC) module is designed to capture both short- and long-range temporal features in multivariate time series through an efficient decomposition of large-kernel convolutions. Instead of directly applying computationally expensive large kernels, TLDC approximates their impact using a sequence of smaller, more efficient convolutions. This approach significantly reduces the computational overhead while preserving the ability to model extensive temporal patterns.

Given the input tensor $\mathbf{X}_{\text{emb}} $ in shape $(M, D, N)$, it is first reshaped to $(M \times D, N)$ to facilitate group convolutions. This grouping structure divides the input into $M \times D$ separate groups, each corresponding to a unique variable-channel pair, as illustrated in ~\autoref{fig:TLDC}. By processing temporal dependencies independently within each group, this configuration allows for parallel computation while preserving the distinct characteristics of each variable.

To approximate a large convolution kernel of size $K$, the module applies a two-stage hierarchical convolutional process. First, a Depth-Wise Convolution (DW Conv) ~\cite{howard2017mobilenets} with a kernel size of $(2d-1)$ is applied, capturing short-term dependencies within a local temporal window:
\vspace{-1pt}
\begin{equation}
    \mathbf{X}_{\text{local}} = \text{DW Conv}(\mathbf{X}_{\text{emb}})
\end{equation}

This operation focuses on adjacent temporal points, enabling the capture of localized patterns while maintaining computational efficiency. The depth-wise approach ensures that each variable-channel pair is processed individually, preserving feature-specific information.

To extend the receptive field, a Depth-Wise Dilated Convolution (DW-D Conv) ~\cite{guo2023visual} is subsequently applied to $\mathbf{X}_{\text{local}}$. Using a kernel size of $\lceil\frac{K}{d}\rceil$ and dilation rate $d$, this convolution broadens the temporal scope, enabling the model to capture long-range dependencies by sampling across wider intervals:
\vspace{-1pt}
\begin{equation}
    \mathbf{X}_{\text{dilated}} = \text{DW-D Conv}(\mathbf{X}_{\text{local}})
\end{equation}

To incorporate both local and distant temporal features, the outputs of the DW Conv and DW-D Conv are combined via element-wise addition:
\vspace{-1pt}
\begin{equation} \mathbf{X}_{\text{combined}} = \mathbf{X}_{\text{dilated}} + \mathbf{X}_{\text{local}} \end{equation}

This summation enables the module to capture information across varying temporal scales, effectively balancing the complexity of large-kernel operations with the ability to extract detailed and broader temporal patterns.

The complexity reduction achieved by the TLDC can be quantified by comparing its parameter count and computational cost in terms of floating-point operations per second (FLOPs), to that of a direct large-kernel convolution. Typically, a single convolution with kernel size $K$ requires:
\vspace{-1pt}
\begin{equation}
    \mathrm{Params}_{{\mathrm{standard}}}=(M\times D)\times(K+1)
\end{equation}
\begin{equation}
    \quad\mathrm{FLOPs}_{{\mathrm{standard}}}=2\times M\times D\times K\times N
\end{equation}

In contrast, the two-stage TLDC requires:
\vspace{-1pt}
\begin{equation}
    \mathrm{Params}_{\mathrm{TLDC}}=M\times D\times(2d+1+\lceil \frac{K}{d}\rceil)
\end{equation}
\begin{equation}
    \mathrm{FLOPs}_{\mathrm{TLDC}}=2\times M\times D\times(2d-1+\lceil \frac{K}{d}\rceil)\times N
\end{equation}

The reduction in complexity can be approximated by the ratios:
\vspace{-1pt}
\begin{equation}
    \frac{\mathrm{Params}_{\mathrm{TLDC}}}{\mathrm{Params}_{\mathrm{standard}}}=\frac{2d+1+\lceil \frac{K}{d}\rceil}{K+1}
\end{equation}
\begin{equation}
    \frac{\mathrm{FLOPs}_{\mathrm{TLDC}}}{\mathrm{FLOPs}_{\mathrm{standard}}}=\frac{2d-1+\lceil \frac{K}{d}\rceil}{K}
\end{equation}

When $d$ is moderately smaller than $K$, these ratios simplify to approximately:
\vspace{-1pt}
\begin{equation}
    \frac{\mathrm{Params}_{\mathrm{TLDC}}}{\mathrm{Params}_{\mathrm{standard}}}\approx\frac{2d}{K}+\frac{1}{d},\quad
    \frac{\mathrm{FLOPs}_{\mathrm{TLDC}}}{\mathrm{FLOPs}_{\mathrm{standard}}}\approx\frac{2d}{K}+\frac{1}{d}
\end{equation}

When $\frac{2d}{K}$ is negligible for $d \ll K$, the dominant term in both cases is $\mathcal{O}(1/d)$. This indicates that the TLDC reduces parameter and computational complexity by a factor inversely proportional to the dilation rate $d$. For instance, with $K=55$ and $d=5$, the parameter reduction factor is approximately 0.39, and the FLOPs reduction factor is around 0.36, leading to a 61\% decrease in parameters and a 64\% decrease in FLOPs.

Through this decomposition of large kernels, the TLDC module effectively captures long-range temporal dependencies while significantly reducing computational cost, enabling efficient temporal feature extraction. This approach is particularly advantageous for data-intensive time series analysis tasks, as it maintains model complexity within practical limits while delivering strong predictive performance and adaptability.

\subsection{Inter-Variable Group Convolution}

\begin{figure}
  \centering
  \includegraphics[width=\linewidth]{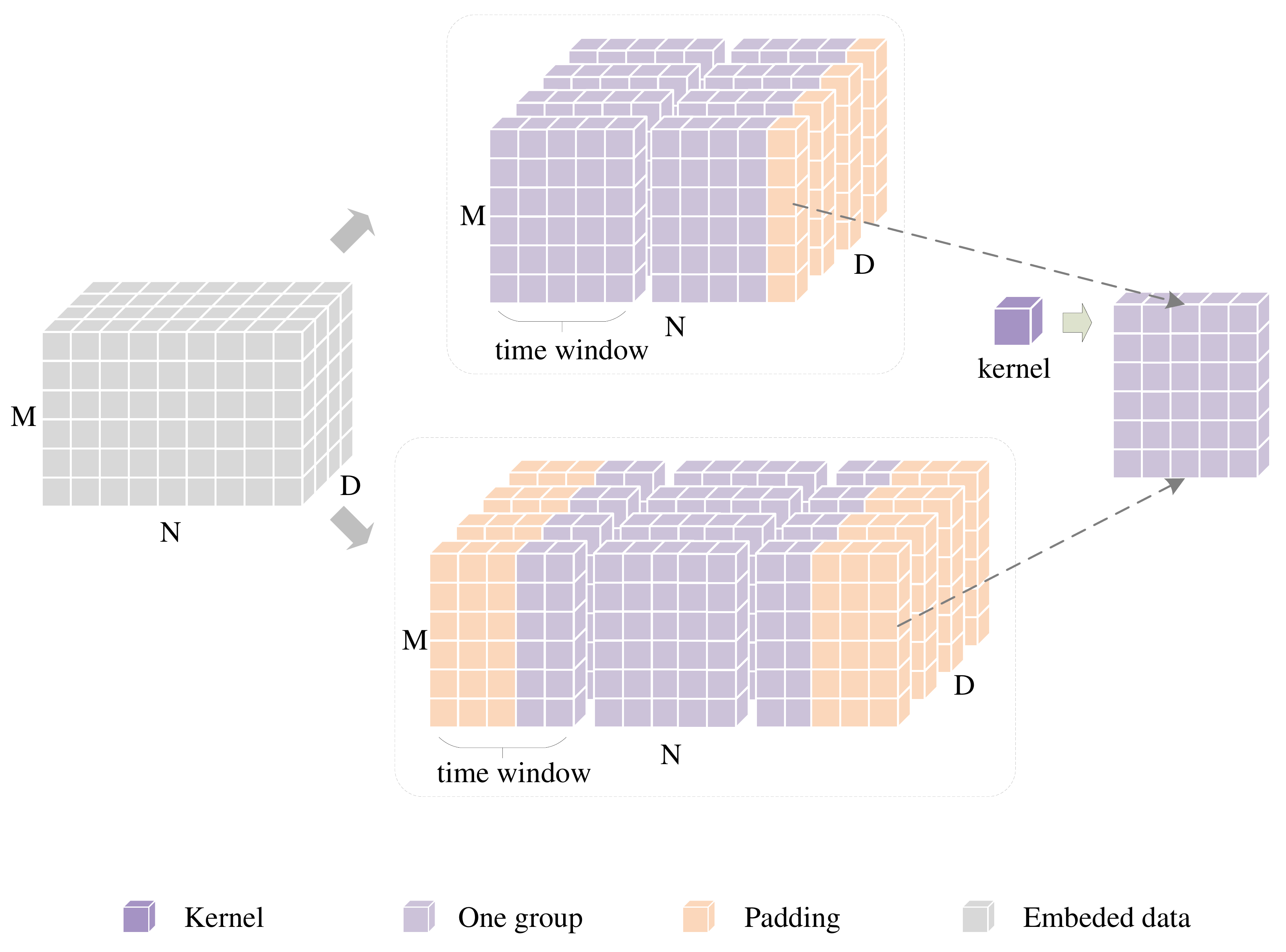}
  \caption{Illustration of the IVGC process. Initially, the data is padded to ensure divisibility by the time window size. Two padding strategies are applied: standard padding (top) and head-tail padding (bottom), creating distinct segmentations of the time dimension. Each group, represented by non-overlapping time windows, undergoes a convolution with a kernel size of 1 to capture inter-variable interactions within localized temporal patches. The outputs from both padding strategies are later aligned, merged, and further processed to produce the final integrated representation.}
  \label{fig:IVGC}
  \vspace{-10pt}
\end{figure}

The Inter-Variable Group Convolution (IVGC) module is designed to efficiently model dependencies among variables in multivariate time series data. By convolving across structured groups of temporal patches, this module captures intricate inter-variable interactions, particularly in scenarios where variables exhibit asynchrony or subtle lags. The group convolution framework focus on local temporal patterns and enhances computational efficiency, making it suitable for data with complex temporal dynamics and variable dependencies.

The input tensor $\mathbf{X}_{\text{combined}} $ in size $(M, D, N)$ is first reshaped into a matrix of size $(N \times M, D)$. To enable group convolutions over multiple time windows, the temporal dimension $N$ is padded to be divisible by the predefined window size $W$. This padding ensures the group convolution to operate on non-overlapping time windows, with the number of groups defined as $\frac{N_{padded}}{W}$, where $N_{padded}$ represents the padded temporal length. Each group spans consecutive time points for all variables, allowing a shared convolutional kernel across the entire window. This configuration captures dynamic inter-variable dependencies within localized windows while preserving computational efficiency.

To ensure consistent alignment of temporal windows across groups, we calculate the necessary padding length, denoted as $N_{\mathrm{pad1}}$, as follows:
\vspace{-1pt}
\begin{equation}
    N_{{\mathrm{pad1}}}=
    \begin{cases}
        0,&\mathrm{if~}N\equiv0\mathrm{~(mod~}W)\\
        W-(N\mathrm{mod~}W),&\mathrm{otherwise}
    \end{cases}
\end{equation}

This formula adjusts the temporal dimension to the nearest multiple of $W$, aligning the time windows for all groups.

To capture a diverse range of dynamic patterns, an additional head-tail padding strategy is applied. This involves shifting the window segmentation by adding padding of $\lfloor\frac{W}{2}\rfloor$ to the start of the temporal dimension and $W - \lfloor\frac{W}{2}\rfloor$ to the end. This configuration slightly offsets the time windows, expanding the convolutional receptive field and allowing the model to capture variable relationships across varying temporal shifts. The additional padding lengths are defined as:
\vspace{-1pt}
\begin{equation}
    \begin{aligned}
        N_{{\mathrm{left\_pad2}}}&=\lfloor\frac{W}{2}\rfloor\\
        N_{{\mathrm{right\_pad2}}}&=W-\lfloor\frac{W}{2}\rfloor+N_{{\mathrm{pad1}}}
    \end{aligned}
\end{equation}

After applying both padding strategies, we obtain two versions of the input tensor: $\mathbf{X}_{\text{padded1}}$ for standard padding and $\mathbf{X}_{\text{padded2}}$ for head-tail padding. Each tensor undergoes a 1-dimensional group convolution along the channel dimension with a kernel size of 1, aggregating inter-variable information within each time window while preserving localized dependencies.

The group convolution for each padded tensor is formulated as follows:
\vspace{-1pt}
\begin{equation}
    \begin{aligned}
        \mathbf{Y}_{{\mathrm{padded1}}}&=\mathrm{Conv}(\mathbf{X}_{{\mathrm{padded1}}})\\
        \mathbf{Y}_{{\mathrm{padded2}}}&=\mathrm{Conv}(\mathbf{X}_{{\mathrm{padded2}}})
    \end{aligned}
\end{equation}

These convolutions are performed separately to capture both standard and shifted temporal patterns. The outputs, $\mathbf{Y}_{\text{padded1}}$ and $\mathbf{Y}_{\text{padded2}}$, are then aligned by removing any extra channels resulting from padding.

Subsequently, the outputs from both convolution paths are merged and processed by a final convolution to further refine the extracted temporal and inter-variable interactions:
\vspace{-1pt}
\begin{equation}
    \mathbf{Y}=\mathrm{Conv}(\mathbf{Y}_{{\mathrm{padded1}}}+\mathbf{Y}_{{\mathrm{padded2}}})
\end{equation}

Finally, the output tensor $\mathbf{Y}$ is reshaped back to its original dimensions $(M, D, N)$, restoring its compatibility with subsequent model layers.

The IVGC module is based on the principle that variables are more likely to exhibit strong interactions within close temporal proximity, rather than over distant time points. By segmenting the data into localized time windows and applying group convolutions, the IVGC captures dynamic inter-variable dependencies that evolve over time, adapting efficiently to these temporal shifts. This design enhances the module's focus on relevant, time-varying relationships while minimizing computational costs by constraining the convolutional operations within compact groups, making it both scalable and effective for complex multivariate time series analysis.

\subsection{Global Temporal-Variable Attention}

\begin{figure}
  \centering
  \includegraphics[width=\linewidth]{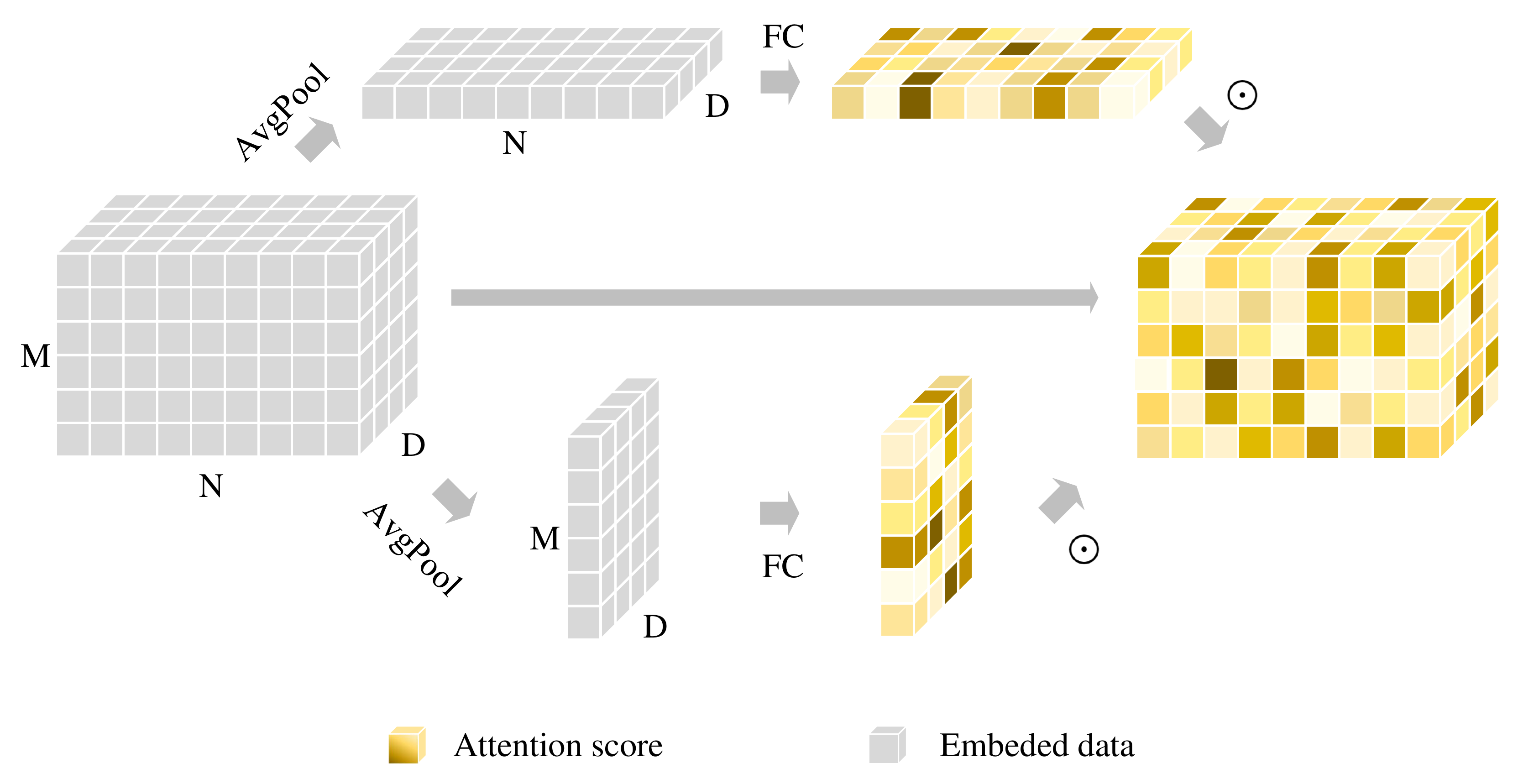}
  \caption{Illustration of the GTVA mechanism using Squeeze-and-Excitation (SE) blocks. Temporal and variable attention pathways separately generate attention weights, which are then multiplied with the convolution outputs to enhance feature learning in both dimensions.}
  \label{fig:GTVA}
  \vspace{-10pt}
\end{figure}

The Global Temporal-Variable Attention (GTVA) mechanism leverages the Squeeze-and-Excitation (SE) principle ~\cite{hu2018squeeze} to selectively enhance feature learning across temporal and variable dimensions in multivariate time series. By decoupling the SE operation into dual-path attention — one focused on temporal dependencies and the other on inter - variable relationships — GTVA extends beyond standard SE applications. This design complements the convolutional layers’ local feature extraction by integrating broader temporal and inter-variable contexts, which is critical for effectively modeling both long-range dependencies and complex variable interactions.

Given the input tensor $\mathbf{Y} \in \mathbb{R}^{M \times D \times N}$, the GTVA module independently generates attention weights for the temporal and variable dimensions. This dual-path approach preserves the unique characteristics of each dimension, avoiding the conflation that may arise from a single-path attention structure. By emphasizing channel-sensitive attention, the mechanism refines feature learning across these axes, further augmenting the model’s ability to capture intricate temporal and inter-variable relationships.

\textbf{Temporal Attention} To capture global temporal dependencies, the input tensor is reshaped to $\mathbf{Y}_{\text{temp}} \in \mathbb{R}^{(N \times D) \times M}$. Global average pooling is then applied along the variable dimension to distill a temporal-centric representation:
\vspace{-1pt}
\begin{equation}
    \mathbf{T}_{{\mathrm{pool}}}=\mathrm{AvgPool}(\mathbf{Y}_{{\mathrm{temp}}})\in\mathbb{R}^{(N\times D)}
\end{equation}

This representation is then processed through a two-layer fully connected (FC) network with a reduction ratio $r$, where $r$ is set to balance computational cost and representation capability:
\vspace{-1pt}
\begin{equation}
    \mathbf{T}_{{\mathrm{atten}}}=\sigma\left(W_{2}\cdot\mathrm{ReLU}(W_{1}\cdot\mathbf{T}_{{\mathrm{pool}}})\right)
\end{equation}

Here, $W_1\in\mathbb{R}^{(N\times D)\times\frac{N\times D}r}$ and $W_2\in\mathbb{R}^{\frac{N\times D}r\times(N\times D)}$ are weight matrices, and $\sigma$ denotes the sigmoid activation function. The resulting attention weights $\mathbf{T}_{{\mathrm{atten}}}$, reshaped back to $(1, D, N)$, provide a temporal attention mask that selectively enhances or suppresses features along the time axis, emphasizing relevant temporal patterns in the input tensor.

\textbf{Variable Attention} In parallel, variable attention captures inter-variable dependencies. The tensor is reshaped to  $\mathbf{Y}_{\text{var}} \in \mathbb{R}^{(M \times D) \times N}$, and global average pooling over the temporal axis produces a variable-centric representation:
\vspace{-1pt}
\begin{equation}
    \mathbf{V}_{{\mathrm{pool}}}=\mathrm{AvgPool}(\mathbf{Y}_{{\mathrm{var}}})\in\mathbb{R}^{(M\times D)}
\end{equation}

This representation is similarly processed through a FC network:
\vspace{-1pt}
\begin{equation}
    \mathbf{V}_{{\mathrm{atten}}}=\sigma\left(W_{4}\cdot\mathrm{ReLU}(W_{3}\cdot\mathbf{V}_{{\mathrm{pool}}})\right)
\end{equation}

where $W_3\in\mathbb{R}^{(M\times D)\times\frac{M\times D}r}$ and $W_4\in\mathbb{R}^{\frac{M\times D}r\times(M\times D)}$. The attention weights $\mathbf{V}_{{\mathrm{atten}}}$, reshaped to $(M, D, 1)$, allow the model to adaptively emphasize or suppress features based on cross-variable dependencies.

The dual attention weights are then combined with the convolution output using the Hadamard product, applying both temporal and variable attention simultaneously:
\vspace{-1pt}
\begin{equation}
    \mathbf{Y}_{{\mathrm{out}}}=\sigma\left(\mathbf{T}_{{\mathrm{atten}}}\odot\mathbf{V}_{{\mathrm{atten}}}\odot\mathbf{Y}\right)
\end{equation}

where $\odot$ denotes the element-wise Hadamard product. This fusion allows the model to emphasize time-sensitive and variable-relevant features concurrently, effectively capturing long-term dependencies and variable-specific interactions.

\textbf{Feedback Integration} To further refine feature representations, the GTVA output $\mathbf{Y}_{\mathrm{out}}$ is incorporated with the block’s input via element-wise multiplication:
\vspace{-1pt}
\begin{equation}
    \mathbf{X}_{\text{emb}}'=\mathbf{Y}_{\mathrm{out}} \odot \mathbf{X}_{\text{emb}}
\end{equation}

Here, $\mathbf{X}_{\text{emb}}'$ is the input of the next Block. This design introduces a feedback mechanism that allows each block’s refined output to dynamically interact with its input, fostering progressive feature enhancement across layers. Unlike conventional residual connections, this feedback strategy cumulatively recalibrates feature representations, continually amplifying relevant information throughout the network, which is particularly beneficial for long-horizon forecasting.

The GTVA module extends traditional SE mechanisms by explicitly considering channel-level interactions within the context of both temporal and variable dimensions. Building on prior convolutions, this design refines forecasting through global temporal and variable attention, enabling precise adjustments at a comprehensive level to capture fine-grained dependencies across the entire time series.

\subsection{Summary}
Algorithm \autoref{alg} outlines the training process of EffiCANet, which incorporates three primary modules within each of the $L$ blocks: TLDC, IVGC, and GTVA. First, the input data $\mathbf{X}$ undergoes patch embedding and is initialized as $\mathbf{Z}^{(0)}$ (lines 4-5). In each block, the temporal features are captured using DW Conv and DW-D Conv in the TLDC module (lines 7-10), while inter-variable dependencies are extracted in the IVGC module through group convolutions with a specified time window size $W$ (lines 11-13). The GTVA module then applies temporal-variable attention pooling, producing an adaptive output that adjusts to both temporal and variable dependencies (lines 14-18). This sequence concludes with an aggregation step, and the final representation $\mathbf{Z}^{(L)}$ is passed to the prediction head for output generation (lines 20). The model parameters are iteratively optimized until convergence by minimizing the prediction loss $\mathcal{L}$ (lines 21-22).

\begin{algorithm}[t!]
  \SetAlgoLined
  \KwIn{Training set $\mathcal{D} = \{(\mathbf{X}, \mathbf{Y})\}$; number of blocks $L$; historical and forecasting horizon $H$, $\tau$; patch size $P$, stride $S$; large-kernel size $K$; dilation rate $d$; time window size $W$; reduction ratio $r$}
  \KwOut{Trained EffiCANet}

  \textbf{Initialize} model with $L$ blocks and patch embedding layer\\

  \Repeat{convergence}{
    Sample $(\mathbf{X}, \mathbf{Y})$ from $\mathcal{D}$\\
    $\mathbf{X}_{\text{emb}} \gets \text{PatchEmbed}(\mathbf{X}; P, S)$\\
    $\mathbf{Z}^{(0)} \gets \mathbf{X}_{\text{emb}}$\\

    \For{$i = 1$ \KwTo $L$}{
      \textbf{// TLDC module}\\
      $\mathbf{X}_{\text{local}} \gets \text{DW Conv}(\mathbf{Z}^{(i-1)})$\\
      $\mathbf{X}_{\text{dilated}} \gets \text{DW-D Conv}(\mathbf{X}_{\text{local}})$\\
      $\mathbf{X}_{\text{combined}} \gets \mathbf{X}_{\text{dilated}} + \mathbf{X}_{\text{local}}$\\

      \textbf{//IVGC module}\\
      Pad $\mathbf{X}_{\text{combined}}$ to match window size $W$\\
      $\mathbf{Y} \gets \text{Group Conv}(\mathbf{X}_{\text{padded}}; W)$\\

      \textbf{// GTVA module}\\
      $\mathbf{T}_{\text{att}} \gets \sigma(W_2 \cdot \text{ReLU}(W_1 \cdot \text{AvgPool}(\mathbf{Y}_{\text{temp}})))$\\
      $\mathbf{V}_{\text{att}} \gets \sigma(W_4 \cdot \text{ReLU}(W_3 \cdot \text{AvgPool}(\mathbf{Y}_{\text{var}})))$\\
      $\mathbf{Y}_{\text{out}} \gets \sigma(\mathbf{T}_{\text{att}} \odot \mathbf{V}_{\text{att}} \odot \mathbf{Y})$\\

      Update $\mathbf{Z}^{(i)} \gets \mathbf{Y}_{\text{out}} \odot \mathbf{Z}^{(i-1)}$\\
    }

    $\hat{\mathbf{Y}} \gets \text{PredictHead}(\mathbf{Z}^{(L)})$\\
    Compute loss $\mathcal{L} \gets \mathcal{L}(\hat{\mathbf{Y}}, \mathbf{Y})$\\
    Update model parameters $\mathbf{\Theta}$ via backpropagation\\
  } 

  \Return Trained EffiCANet\\
  \caption{Training of EffiCANet}
  \label{alg}
\end{algorithm}
\vspace{-6pt}

\section{Experiments}\label{Experiments}

\subsection{Experimental Setup}
\subsubsection{Datasets and Evaluation Metrics}
We evaluate EffiCANet's performance across nine widely-adopted multivariate time series datasets.

\begin{itemize}[leftmargin=*, labelwidth=\parindent]
\item\textbf{ETT:} The Electricity Transformer Temperature (ETT) dataset ~\cite{zhou2021informer} contains transformer oil temperature data, including four subsets — ETTm1, ETTm2, ETTh1, and ETTh2, each representing different temporal granularities and historical spans.

\item\textbf{Electricity:} The Electricity dataset ~\cite{wu2021autoformer} consists of hourly electricity consumption data from 321 households.

\item\textbf{Weather:} The Weather dataset ~\cite{wu2021autoformer} includes meteorological variables such as temperature, humidity, wind speed, and pressure, collected from multiple weather stations.

\item\textbf{Traffic:} The Traffic dataset ~\cite{wu2021autoformer} contains hourly occupancy rates of roadways in the San Francisco Bay Area.

\item\textbf{Exchange:} The Exchange dataset ~\cite{lai2018modeling} records daily exchange rates of eight major world currencies. 

\item\textbf{ILI:} The Influenza-like Illness (ILI) dataset ~\cite{wu2021autoformer} tracks the weekly percentage of doctor visits for influenza-like illnesses in the United States.
\end{itemize}

Table \ref{datasets} provides a detailed overview of each dataset, including the number of timesteps, features, and sampling frequency.

\begin{table}[th]
\caption{Summary of each dataset.}
\label{datasets}
\centering
\renewcommand{\multirowsetup}{\centering}
\setlength{\tabcolsep}{7.5pt}
\scalebox{1}{
\begin{tabular}{cccc}
\toprule
\scalebox{0.9}{Datasets} & \scalebox{0.9}{Timesteps} & \scalebox{0.9}{Features} & \scalebox{0.9}{Frequency}\\
\midrule
\scalebox{0.9}{ETTh1} & \scalebox{0.9}{17420} & \scalebox{0.9}{7} & \scalebox{0.9}{Hourly}\\
\scalebox{0.9}{ETTh2} & \scalebox{0.9}{17420} & \scalebox{0.9}{7} & \scalebox{0.9}{Hourly}\\
\scalebox{0.9}{ETTm1} & \scalebox{0.9}{69680} & \scalebox{0.9}{7} & \scalebox{0.9}{15 mins}\\
\scalebox{0.9}{ETTm2} & \scalebox{0.9}{69680} & \scalebox{0.9}{7} & \scalebox{0.9}{15 mins}\\
\scalebox{0.9}{Electricity} & \scalebox{0.9}{26304} & \scalebox{0.9}{321} & \scalebox{0.9}{Hourly}\\
\scalebox{0.9}{Weather} & \scalebox{0.9}{52696} & \scalebox{0.9}{21} & \scalebox{0.9}{10 mins}\\
\scalebox{0.9}{Traffic} & \scalebox{0.9}{17544} & \scalebox{0.9}{862} & \scalebox{0.9}{Hourly}\\
\scalebox{0.9}{Exchange} & \scalebox{0.9}{7207} & \scalebox{0.9}{8} & \scalebox{0.9}{Daily}\\
\scalebox{0.9}{ILI} & \scalebox{0.9}{966} & \scalebox{0.9}{7} & \scalebox{0.9}{Weekly}\\   
\bottomrule
\end{tabular}}
\end{table}

For evaluation, we use Mean Squared Error (MSE) and Mean Absolute Error (MAE) as performance metrics, where lower values reflect better forecasting accuracy.

\subsubsection{Baselines}

We evaluate our model against a range of baselines across three categories: convolution-based, Transformer-based, and MLP-based models. 

\begin{itemize}[leftmargin=*, labelwidth=\parindent]

\item\textbf{Convolution-based Baselines.} ConvTimeNet ~\cite{cheng2024convtimenet} utilizes deepwise and pointwise convolutions to capture both global sequence and cross-variable dependence. ModernTCN ~\cite{luo2024moderntcn} modernizes the traditional TCN by expanding the receptive field and offers a pure convolution structure for time series analysis. TimesNet ~\cite{wutimesnet} focuses on multi-scale temporal feature extraction through specialized convolutions, while MICN ~\cite{wang2023micn} integrates information across various time scales for improved prediction accuracy. 

\item\textbf{Transformer-based Baselines.} PatchTST ~\cite{nietime}
introduces a patching mechanism with channel-independence, and Crossformer ~\cite{zhang2023crossformer} applies cross-dimensional attention to capture both temporal and inter-variable dependencies. 

\item\textbf{MLP-based Baselines.} MTS-mixer ~\cite{li2023mts} replaces attention mechanisms with factorized MLP modules to separately model temporal and feature dependencies. DLinear ~\cite{zeng2023transformers}, RLinear, RMLP ~\cite{li2023revisiting} demonstrate the effectiveness of simple linear layers and MLP structures in long-term forecasting.
\end{itemize}

\subsubsection{Implementation Details}
The experiments were implemented on an NVIDIA GeForce RTX 4090 GPU. We used the Adam optimizer with learning rates ranging from $1e^{-4}$ to $1e^{-2}$, adjusting for each dataset. The batch sizes were set to 32, 128, 256, or 512 depending on the dataset size. The model consisted of 1 to 3 blocks, with training running up to 100 epochs and early stopping triggered if validation loss stagnated for 20 epochs. A channel dimension size of 64 was employed consistently throughout all experiments. The simulated temporal large-kernel convolution size was set to 55 with a dilation rate of 5. Input sequence lengths were optimized per dataset, with values of 128, 336, 512, 720, and 960. Baselines followed the original configurations to ensure fair comparisons.

\subsection{Main Results}

\begin{table*}[htbp]
\setlength{\abovecaptionskip}{0.cm}
\setlength{\belowcaptionskip}{-0.cm}
  \centering
  \small
  \caption{Performance comparison of our method against various baseline models across different datasets and prediction lengths. \textcolor[rgb]{ 1,  0,  0}{\textbf{Red}} indicates the best performance, and \textcolor[rgb]{ 0,  0,  1}{\underline{blue}} indicates the second-best performance.}
  \setlength{\tabcolsep}{1.5pt}
    \begin{tabular}{cccccccccccccccccccccccc}
    \toprule
    \multicolumn{2}{c}{Models} & \multicolumn{2}{c}{EffiCANet} & \multicolumn{2}{c}{\begin{tabular}[c]{@{}c@{}}\footnotesize ConvTimeNet\\      (2024)\end{tabular}} & \multicolumn{2}{c}{\begin{tabular}[c]{@{}c@{}} ModernTCN\\      (2024)\end{tabular}} & \multicolumn{2}{c}{\begin{tabular}[c]{@{}c@{}}PatchTST\\      (2023)\end{tabular}} & \multicolumn{2}{c}{\begin{tabular}[c]{@{}c@{}}Crossformer\\      (2023)\end{tabular}} & \multicolumn{2}{c}{\begin{tabular}[c]{@{}c@{}}MTS-mixer\\      (2023)\end{tabular}} & \multicolumn{2}{c}{\begin{tabular}[c]{@{}c@{}}DLinear\\      (2023)\end{tabular}} & \multicolumn{2}{c}{\begin{tabular}[c]{@{}c@{}}TimesNet\\      (2023)\end{tabular}} & \multicolumn{2}{c}{\begin{tabular}[c]{@{}c@{}}MICN\\      (2023)\end{tabular}} & \multicolumn{2}{c}{\begin{tabular}[c]{@{}c@{}}RLinear\\      (2023)\end{tabular}} & \multicolumn{2}{c}{\begin{tabular}[c]{@{}c@{}}RMLP\\      (2023)\end{tabular}} \\
\cmidrule{3-24}    \multicolumn{2}{c}{Metric} & MSE & MAE & MSE & MAE & MSE & MAE & MSE & MAE & MSE & MAE & MSE & MAE & MSE & MAE & MSE & MAE & MSE & MAE & MSE & MAE & MSE & MAE \\
    \midrule
    \multirow{5}[4]{*}{\begin{sideways}ETTm1\end{sideways}} & 96 & \textcolor[rgb]{ 1,  0,  0}{\textbf{0.288 }} & \textcolor[rgb]{ 1,  0,  0}{\textbf{0.337 }} & 0.292  & 0.345  & 0.292  & 0.346  & \textcolor[rgb]{ 0,  0,  1}{\underline{0.290} } & \textcolor[rgb]{ 0,  0,  1}{\underline{0.342} } & 0.316  & 0.373  & 0.314  & 0.358  & 0.299  & 0.343  & 0.338  & 0.375  & 0.314  & 0.360  & 0.301  & \textcolor[rgb]{ 0,  0,  1}{\underline{0.342} } & 0.298  & 0.345  \\
       & 192 & \textcolor[rgb]{ 0,  0,  1}{\underline{0.331} } & \textcolor[rgb]{ 0,  0,  1}{\underline{0.364} } & \textcolor[rgb]{ 1,  0,  0}{\textbf{0.329 }} & 0.368  & 0.332  & 0.368  & 0.332  & 0.369  & 0.377  & 0.411  & 0.354  & 0.386  & 0.335  & 0.365  & 0.371  & 0.387  & 0.359  & 0.387  & 0.335  & \textcolor[rgb]{ 1,  0,  0}{\textbf{0.363 }} & 0.344  & 0.375  \\
       & 336 & \textcolor[rgb]{ 1,  0,  0}{\textbf{0.362 }} & \textcolor[rgb]{ 0,  0,  1}{\underline{0.384} } & \textcolor[rgb]{ 0,  0,  1}{\underline{0.363} } & 0.390  & 0.365  & 0.391  & 0.366  & 0.392  & 0.431  & 0.442  & 0.384  & 0.405  & 0.369  & 0.386  & 0.410  & 0.411  & 0.398  & 0.413  & 0.370  & \textcolor[rgb]{ 1,  0,  0}{\textbf{0.383 }} & 0.390  & 0.410  \\
       & 720 & \textcolor[rgb]{ 1,  0,  0}{\textbf{0.407 }} & \textcolor[rgb]{ 1,  0,  0}{\textbf{0.413 }} & 0.427  & 0.428  & 0.416  & \textcolor[rgb]{ 0,  0,  1}{\underline{0.417} } & 0.416  & 0.420  & 0.600  & 0.547  & 0.427  & 0.432  & 0.425  & 0.421  & 0.478  & 0.450  & 0.459  & 0.464  & 0.425  & \textcolor[rgb]{ 0,  0,  1}{\underline{0.414} } & 0.445  & 0.441  \\
\cmidrule{2-24}       & Avg & \textcolor[rgb]{ 1,  0,  0}{\textbf{0.347 }} & \textcolor[rgb]{ 1,  0,  0}{\textbf{0.375 }} & 0.353  & 0.383  & \textcolor[rgb]{ 0,  0,  1}{\underline{0.351} } & 0.381  & \textcolor[rgb]{ 0,  0,  1}{\underline{0.351} } & 0.381  & 0.431  & 0.443  & 0.370  & 0.395  & 0.357  & 0.379  & 0.400  & 0.406  & 0.383  & 0.406  & 0.358  & \textcolor[rgb]{ 0,  0,  1}{\underline{0.376} } & 0.369  & 0.393  \\
    \midrule
    \multirow{5}[4]{*}{\begin{sideways}ETTm2\end{sideways}} & 96 & \textcolor[rgb]{ 0,  0,  1}{\underline{0.165} } & \textcolor[rgb]{ 0,  0,  1}{\underline{0.254} } & 0.167  & 0.257  & 0.166  & 0.256  & \textcolor[rgb]{ 0,  0,  1}{\underline{0.165} } & 0.255  & 0.296  & 0.352  & 0.177  & 0.259  & 0.167  & 0.260  & 0.187  & 0.267  & 0.178  & 0.273  & \textcolor[rgb]{ 1,  0,  0}{\textbf{0.164 }} & \textcolor[rgb]{ 1,  0,  0}{\textbf{0.253 }} & 0.174  & 0.259  \\
       & 192 & \textcolor[rgb]{ 1,  0,  0}{\textbf{0.217 }} & \textcolor[rgb]{ 1,  0,  0}{\textbf{0.289 }} & 0.222  & 0.295  & 0.222  & 0.293  & 0.220  & 0.292  & 0.342  & 0.385  & 0.241  & 0.303  & 0.224  & 0.303  & 0.249  & 0.309  & 0.245  & 0.316  & \textcolor[rgb]{ 0,  0,  1}{\underline{0.219} } & \textcolor[rgb]{ 0,  0,  1}{\underline{0.290} } & 0.236  & 0.303  \\
       & 336 & \textcolor[rgb]{ 1,  0,  0}{\textbf{0.271 }} & 0.327  & 0.276  & 0.329  & \textcolor[rgb]{ 0,  0,  1}{\underline{0.272} } & \textcolor[rgb]{ 1,  0,  0}{\textbf{0.324 }} & 0.274  & 0.329  & 0.410  & 0.425  & 0.297  & 0.338  & 0.281  & 0.342  & 0.321  & 0.351  & 0.295  & 0.350  & 0.273  & \textcolor[rgb]{ 0,  0,  1}{\underline{0.326} } & 0.291  & 0.338  \\
       & 720 & \textcolor[rgb]{ 1,  0,  0}{\textbf{0.340 }} & \textcolor[rgb]{ 1,  0,  0}{\textbf{0.377 }} & 0.358  & \textcolor[rgb]{ 0,  0,  1}{\underline{0.381} } & \textcolor[rgb]{ 0,  0,  1}{\underline{0.351} } & \textcolor[rgb]{ 0,  0,  1}{\underline{0.381} } & 0.362  & 0.385  & 0.563  & 0.538  & 0.396  & 0.398  & 0.397  & 0.421  & 0.497  & 0.403  & 0.389  & 0.406  & 0.366  & 0.385  & 0.371  & 0.391  \\
\cmidrule{2-24}       & Avg & \textcolor[rgb]{ 1,  0,  0}{\textbf{0.248 }} & \textcolor[rgb]{ 1,  0,  0}{\textbf{0.312 }} & 0.256  & 0.316  & \textcolor[rgb]{ 0,  0,  1}{\underline{0.253} } & \textcolor[rgb]{ 0,  0,  1}{\underline{0.314} } & 0.255  & 0.315  & 0.402  & 0.425  & 0.277  & 0.325  & 0.267  & 0.332  & 0.291  & 0.333  & 0.277  & 0.336  & 0.256  & \textcolor[rgb]{ 0,  0,  1}{\underline{0.314} } & 0.268  & 0.322  \\
    \midrule
    \multirow{5}[4]{*}{\begin{sideways}ETTh1\end{sideways}} & 96 & \textcolor[rgb]{ 1,  0,  0}{\textbf{0.361 }} & \textcolor[rgb]{ 1,  0,  0}{\textbf{0.391 }} & 0.368  & 0.394  & 0.368  & 0.394  & 0.370  & 0.399  & 0.386  & 0.429  & 0.372  & 0.395  & 0.375  & 0.399  & 0.384  & 0.402  & 0.396  & 0.427  & \textcolor[rgb]{ 0,  0,  1}{\underline{0.366} } & \textcolor[rgb]{ 1,  0,  0}{\textbf{0.391 }} & 0.390  & 0.410  \\
       & 192 & \textcolor[rgb]{ 1,  0,  0}{\textbf{0.399 }} & \textcolor[rgb]{ 1,  0,  0}{\textbf{0.412 }} & 0.406  & 0.414  & 0.405  & 0.413  & 0.413  & 0.421  & 0.419  & 0.444  & 0.416  & 0.426  & 0.405  & 0.416  & 0.557  & 0.436  & 0.430  & 0.453  & \textcolor[rgb]{ 0,  0,  1}{\underline{0.404} } & \textcolor[rgb]{ 1,  0,  0}{\textbf{0.412 }} & 0.430  & 0.432  \\
       & 336 & \textcolor[rgb]{ 1,  0,  0}{\textbf{0.374 }} & \textcolor[rgb]{ 1,  0,  0}{\textbf{0.407 }} & 0.405  & 0.420  & \textcolor[rgb]{ 0,  0,  1}{\underline{0.391} } & \textcolor[rgb]{ 0,  0,  1}{\underline{0.412} } & 0.422  & 0.436  & 0.440  & 0.461  & 0.455  & 0.449  & 0.439  & 0.443  & 0.491  & 0.469  & 0.433  & 0.458  & 0.420  & 0.423  & 0.441  & 0.441  \\
       & 720 & \textcolor[rgb]{ 1,  0,  0}{\textbf{0.415 }} & \textcolor[rgb]{ 1,  0,  0}{\textbf{0.444 }} & \textcolor[rgb]{ 0,  0,  1}{\underline{0.442} } & 0.457  & 0.450  & 0.461  & 0.447  & 0.466  & 0.519  & 0.524  & 0.475  & 0.472  & 0.472  & 0.490  & 0.521  & 0.500  & 0.474  & 0.508  & \textcolor[rgb]{ 0,  0,  1}{\underline{0.442} } & \textcolor[rgb]{ 0,  0,  1}{\underline{0.456} } & 0.506  & 0.495  \\
\cmidrule{2-24}       & Avg & \textcolor[rgb]{ 1,  0,  0}{\textbf{0.387 }} & \textcolor[rgb]{ 1,  0,  0}{\textbf{0.414 }} & 0.405  & 0.421  & \textcolor[rgb]{ 0,  0,  1}{\underline{0.404} } & \textcolor[rgb]{ 0,  0,  1}{\underline{0.420} } & 0.413  & 0.431  & 0.441  & 0.465  & 0.430  & 0.436  & 0.423  & 0.437  & 0.458  & 0.450  & 0.433  & 0.462  & 0.408  & 0.421  & 0.442  & 0.445  \\
    \midrule
    \multirow{5}[4]{*}{\begin{sideways}ETTh2\end{sideways}} & 96 & \textcolor[rgb]{ 1,  0,  0}{\textbf{0.251 }} & \textcolor[rgb]{ 1,  0,  0}{\textbf{0.323 }} & 0.264  & \textcolor[rgb]{ 0,  0,  1}{\underline{0.330} } & 0.263  & 0.332  & 0.274  & 0.336  & 0.383  & 0.420  & 0.307  & 0.354  & 0.289  & 0.353  & 0.340  & 0.374  & 0.289  & 0.357  & \textcolor[rgb]{ 0,  0,  1}{\underline{0.262} } & 0.331  & 0.288  & 0.352  \\
       & 192 & \textcolor[rgb]{ 1,  0,  0}{\textbf{0.299 }} & \textcolor[rgb]{ 1,  0,  0}{\textbf{0.362 }} & \textcolor[rgb]{ 0,  0,  1}{\underline{0.316} } & \textcolor[rgb]{ 0,  0,  1}{\underline{0.368} } & 0.320  & 0.374  & 0.339  & 0.379  & 0.421  & 0.450  & 0.374  & 0.399  & 0.383  & 0.418  & 0.402  & 0.414  & 0.409  & 0.438  & 0.320  & 0.374  & 0.343  & 0.387  \\
       & 336 & \textcolor[rgb]{ 1,  0,  0}{\textbf{0.295 }} & \textcolor[rgb]{ 1,  0,  0}{\textbf{0.359 }} & 0.315  & 0.378  & \textcolor[rgb]{ 0,  0,  1}{\underline{0.313} } & \textcolor[rgb]{ 0,  0,  1}{\underline{0.376} } & 0.329  & 0.380  & 0.449  & 0.459  & 0.398  & 0.432  & 0.448  & 0.465  & 0.452  & 0.452  & 0.417  & 0.452  & 0.325  & 0.386  & 0.353  & 0.402  \\
       & 720 & \textcolor[rgb]{ 1,  0,  0}{\textbf{0.372 }} & \textcolor[rgb]{ 1,  0,  0}{\textbf{0.419 }} & 0.382  & 0.425  & 0.392  & 0.433  & 0.379  & 0.422  & 0.472  & 0.497  & 0.463  & 0.465  & 0.605  & 0.551  & 0.462  & 0.468  & 0.426  & 0.473  & \textcolor[rgb]{ 1,  0,  0}{\textbf{0.372 }} & \textcolor[rgb]{ 0,  0,  1}{\underline{0.421} } & 0.410  & 0.440  \\
\cmidrule{2-24}       & Avg & \textcolor[rgb]{ 1,  0,  0}{\textbf{0.304 }} & \textcolor[rgb]{ 1,  0,  0}{\textbf{0.366 }} & \textcolor[rgb]{ 0,  0,  1}{\underline{0.319} } & \textcolor[rgb]{ 0,  0,  1}{\underline{0.375} } & 0.322  & 0.379  & 0.330  & 0.379  & 0.431  & 0.457  & 0.386  & 0.413  & 0.431  & 0.447  & 0.414  & 0.427  & 0.385  & 0.430  & 0.320  & 0.378  & 0.349  & 0.395  \\
    \midrule
    \multirow{5}[4]{*}{\begin{sideways}Electricity\end{sideways}} & 96 & \textcolor[rgb]{ 1,  0,  0}{\textbf{0.129 }}  & 0.226  & 0.132  & 0.226  & \textcolor[rgb]{ 1,  0,  0}{\textbf{0.129 }} & 0.226  & \textcolor[rgb]{ 1,  0,  0}{\textbf{0.129 }} & \textcolor[rgb]{ 1,  0,  0}{\textbf{0.222 }} & 0.187  & 0.283  & 0.141  & 0.243  & 0.153  & 0.237  & 0.168  & 0.272  & 0.159  & 0.267  & 0.140  & 0.235  & \textcolor[rgb]{ 1,  0,  0}{\textbf{0.129 }} & \textcolor[rgb]{ 0,  0,  1}{\underline{0.224} } \\
       & 192 & \textcolor[rgb]{ 0,  0,  1}{\underline{0.145} } & 0.241  & 0.148  & 0.241  & \textcolor[rgb]{ 1,  0,  0}{\textbf{0.143 }} & \textcolor[rgb]{ 1,  0,  0}{\textbf{0.239 }} & 0.147  & \textcolor[rgb]{ 0,  0,  1}{\underline{0.240} } & 0.258  & 0.330  & 0.163  & 0.261  & 0.152  & 0.249  & 0.184  & 0.289  & 0.168  & 0.279  & 0.154  & 0.248  & 0.147  & \textcolor[rgb]{ 0,  0,  1}{\underline{0.240} } \\
       & 336 & \textcolor[rgb]{ 1,  0,  0}{\textbf{0.160 }} & \textcolor[rgb]{ 1,  0,  0}{\textbf{0.257 }} & 0.165  & 0.259  & \textcolor[rgb]{ 0,  0,  1}{\underline{0.161} } & 0.259  & 0.163  & 0.259  & 0.323  & 0.369  & 0.176  & 0.277  & 0.169  & 0.267  & 0.198  & 0.300  & 0.196  & 0.308  & 0.171  & 0.264  & 0.164  & \textcolor[rgb]{ 1,  0,  0}{\textbf{0.257 }} \\
       & 720 & \textcolor[rgb]{ 1,  0,  0}{\textbf{0.187 }} & \textcolor[rgb]{ 1,  0,  0}{\textbf{0.285 }} & 0.205  & 0.293  & \textcolor[rgb]{ 0,  0,  1}{\underline{0.191} } & \textcolor[rgb]{ 0,  0,  1}{\underline{0.286} } & 0.197  & 0.290  & 0.404  & 0.423  & 0.212  & 0.308  & 0.233  & 0.344  & 0.220  & 0.320  & 0.203  & 0.312  & 0.209  & 0.297  & 0.203  & 0.291  \\
\cmidrule{2-24}       & Avg & \textcolor[rgb]{ 1,  0,  0}{\textbf{0.156 }} & \textcolor[rgb]{ 1,  0,  0}{\textbf{0.252 }} & 0.163  & 0.255  & \textcolor[rgb]{ 1,  0,  0}{\textbf{0.156 }} & \textcolor[rgb]{ 0,  0,  1}{\underline{0.253} } & 0.159  & 0.253  & 0.293  & 0.351  & 0.173  & 0.272  & 0.177  & 0.274  & 0.192  & 0.295  & 0.182  & 0.292  & 0.169  & 0.261  & 0.161  & \textcolor[rgb]{ 0,  0,  1}{\underline{0.253} } \\
    \midrule
    \multirow{5}[4]{*}{\begin{sideways}Weather\end{sideways}} & 96 & \textcolor[rgb]{ 1,  0,  0}{\textbf{0.145 }} & \textcolor[rgb]{ 1,  0,  0}{\textbf{0.197 }} & 0.155  & 0.205  & \textcolor[rgb]{ 0,  0,  1}{\underline{0.149} } & 0.200  & \textcolor[rgb]{ 0,  0,  1}{\underline{0.149} } & \textcolor[rgb]{ 0,  0,  1}{\underline{0.198} } & 0.153  & 0.217  & 0.156  & 0.206  & 0.152  & 0.237  & 0.172  & 0.220  & 0.161  & 0.226  & 0.175  & 0.225  & \textcolor[rgb]{ 0,  0,  1}{\underline{0.149} } & 0.202  \\
       & 192 & \textcolor[rgb]{ 1,  0,  0}{\textbf{0.189 }} & \textcolor[rgb]{ 1,  0,  0}{\textbf{0.239 }} & 0.200  & 0.249  & 0.196  & 0.245  & \textcolor[rgb]{ 0,  0,  1}{\underline{0.194} } & \textcolor[rgb]{ 0,  0,  1}{\underline{0.241} } & 0.197  & 0.269  & 0.199  & 0.248  & 0.220  & 0.282  & 0.219  & 0.261  & 0.220  & 0.283  & 0.218  & 0.260  & \textcolor[rgb]{ 0,  0,  1}{\underline{0.194} } & 0.242  \\
       & 336 & \textcolor[rgb]{ 1,  0,  0}{\textbf{0.234 }} & \textcolor[rgb]{ 0,  0,  1}{\underline{0.279} } & 0.252  & 0.287  & \textcolor[rgb]{ 0,  0,  1}{\underline{0.238} } & \textcolor[rgb]{ 1,  0,  0}{\textbf{0.277 }} & 0.245  & 0.282  & 0.252  & 0.311  & 0.249  & 0.291  & 0.265  & 0.319  & 0.280  & 0.306  & 0.275  & 0.328  & 0.265  & 0.294  & 0.243  & 0.282  \\
       & 720 & \textcolor[rgb]{ 1,  0,  0}{\textbf{0.309 }} & \textcolor[rgb]{ 1,  0,  0}{\textbf{0.331 }} & 0.321  & 0.335  & 0.314  & 0.334  & 0.314  & 0.334  & 0.318  & 0.363  & 0.336  & 0.343  & 0.323  & 0.362  & 0.365  & 0.359  & \textcolor[rgb]{ 0,  0,  1}{\underline{0.311} } & 0.356  & 0.329  & 0.339  & 0.316  & \textcolor[rgb]{ 0,  0,  1}{\underline{0.333} } \\
\cmidrule{2-24}       & Avg & \textcolor[rgb]{ 1,  0,  0}{\textbf{0.219 }} & \textcolor[rgb]{ 1,  0,  0}{\textbf{0.262 }} & 0.232  & 0.269  & \textcolor[rgb]{ 0,  0,  1}{\underline{0.224} } & \textcolor[rgb]{ 0,  0,  1}{\underline{0.264} } & 0.226  & \textcolor[rgb]{ 0,  0,  1}{\underline{0.264} } & 0.230  & 0.290  & 0.235  & 0.272  & 0.240  & 0.300  & 0.259  & 0.287  & 0.242  & 0.298  & 0.247  & 0.279  & 0.225  & 0.265  \\
    \midrule
    \multirow{5}[4]{*}{\begin{sideways}Traffic\end{sideways}} & 96 & \textcolor[rgb]{ 1,  0,  0}{\textbf{0.354 }} & 0.261  & 0.376  & 0.265  & 0.368  & \textcolor[rgb]{ 0,  0,  1}{\underline{0.253} } & \textcolor[rgb]{ 0,  0,  1}{\underline{0.360} } & \textcolor[rgb]{ 1,  0,  0}{\textbf{0.249 }} & 0.512  & 0.290  & 0.462  & 0.332  & 0.410  & 0.282  & 0.593  & 0.321  & 0.508  & 0.301  & 0.496  & 0.375  & 0.430  & 0.327  \\
       & 192 & \textcolor[rgb]{ 1,  0,  0}{\textbf{0.371 }} & 0.269  & 0.392  & 0.271  & \textcolor[rgb]{ 0,  0,  1}{\underline{0.379} } & \textcolor[rgb]{ 0,  0,  1}{\underline{0.261} } & \textcolor[rgb]{ 0,  0,  1}{\underline{0.379} } & \textcolor[rgb]{ 1,  0,  0}{\textbf{0.256 }} & 0.523  & 0.297  & 0.488  & 0.354  & 0.423  & 0.287  & 0.617  & 0.336  & 0.536  & 0.315  & 0.503  & 0.377  & 0.451  & 0.340  \\
       & 336 & \textcolor[rgb]{ 1,  0,  0}{\textbf{0.388 }} & 0.279  & 0.405  & 0.277  & 0.397  & \textcolor[rgb]{ 0,  0,  1}{\underline{0.270} } & \textcolor[rgb]{ 0,  0,  1}{\underline{0.392} } & \textcolor[rgb]{ 1,  0,  0}{\textbf{0.264 }} & 0.530  & 0.300  & 0.498  & 0.360  & 0.436  & 0.296  & 0.629  & 0.336  & 0.525  & 0.310  & 0.517  & 0.382  & 0.470  & 0.351  \\
       & 720 & \textcolor[rgb]{ 0,  0,  1}{\underline{0.435} } & 0.301  & 0.436  & \textcolor[rgb]{ 0,  0,  1}{\underline{0.294} } & 0.440  & 0.296  & \textcolor[rgb]{ 1,  0,  0}{\textbf{0.432 }} & \textcolor[rgb]{ 1,  0,  0}{\textbf{0.286 }} & 0.573  & 0.313  & 0.529  & 0.370  & 0.466  & 0.315  & 0.640  & 0.350  & 0.571  & 0.323  & 0.555  & 0.398  & 0.513  & 0.372  \\
\cmidrule{2-24}       & Avg & \textcolor[rgb]{ 1,  0,  0}{\textbf{0.387 }} & 0.278  & 0.402  & 0.277  & 0.396  & \textcolor[rgb]{ 0,  0,  1}{\underline{0.270} } & \textcolor[rgb]{ 0,  0,  1}{\underline{0.391} } & \textcolor[rgb]{ 1,  0,  0}{\textbf{0.264 }} & 0.535  & 0.300  & 0.494  & 0.354  & 0.434  & 0.295  & 0.620  & 0.336  & 0.535  & 0.312  & 0.518  & 0.383  & 0.466  & 0.348  \\
    \midrule
    \multirow{5}[4]{*}{\begin{sideways}Exchange\end{sideways}} & 96 & \textcolor[rgb]{ 1,  0,  0}{\textbf{0.080 }} & \textcolor[rgb]{ 1,  0,  0}{\textbf{0.196 }} & 0.086  & 0.204  & \textcolor[rgb]{ 1,  0,  0}{\textbf{0.080 }} & \textcolor[rgb]{ 1,  0,  0}{\textbf{0.196 }} & 0.093  & 0.214  & 0.186  & 0.346  & 0.083  & 0.201  & 0.081  & 0.203  & 0.107  & 0.234  & 0.102  & 0.235  & 0.083  & 0.201  & 0.083  & 0.201  \\
       & 192 & \textcolor[rgb]{ 0,  0,  1}{\underline{0.166} } & \textcolor[rgb]{ 1,  0,  0}{\textbf{0.288 }} & 0.184  & 0.303  & \textcolor[rgb]{ 0,  0,  1}{\underline{0.166} } & \textcolor[rgb]{ 1,  0,  0}{\textbf{0.288 }} & 0.192  & 0.312  & 0.467  & 0.522  & 0.174  & 0.296  & \textcolor[rgb]{ 1,  0,  0}{\textbf{0.157 }} & 0.293  & 0.226  & 0.344  & 0.172  & 0.316  & 0.170  & 0.293  & 0.170  & 0.292  \\
       & 336 & \textcolor[rgb]{ 0,  0,  1}{\underline{0.305} } & \textcolor[rgb]{ 1,  0,  0}{\textbf{0.396 }} & 0.341  & 0.421  & 0.307  & \textcolor[rgb]{ 0,  0,  1}{\underline{0.398} } & 0.350  & 0.432  & 0.783  & 0.721  & 0.336  & 0.417  & \textcolor[rgb]{ 0,  0,  1}{\underline{0.305} } & 0.414  & 0.367  & 0.448  & \textcolor[rgb]{ 1,  0,  0}{\textbf{0.272 }} & 0.407  & 0.309  & 0.401  & 0.309  & 0.401  \\
       & 720 & \textcolor[rgb]{ 1,  0,  0}{\textbf{0.636 }} & \textcolor[rgb]{ 1,  0,  0}{\textbf{0.569 }} & 0.879  & 0.701  & 0.656  & \textcolor[rgb]{ 0,  0,  1}{\underline{0.582} } & 0.911  & 0.716  & 1.367  & 0.943  & 0.900  & 0.715  & \textcolor[rgb]{ 0,  0,  1}{\underline{0.643} } & 0.601  & 0.964  & 0.746  & 0.714  & 0.658  & 0.817  & 0.680  & 0.816  & 0.680  \\
\cmidrule{2-24}       & Avg & \textcolor[rgb]{ 1,  0,  0}{\textbf{0.297 }} & \textcolor[rgb]{ 1,  0,  0}{\textbf{0.362 }} & 0.373  & 0.407  & 0.302  & \textcolor[rgb]{ 0,  0,  1}{\underline{0.366} } & 0.387  & 0.419  & 0.701  & 0.633  & 0.373  & 0.407  & \textcolor[rgb]{ 1,  0,  0}{\textbf{0.297 }} & 0.378  & 0.416  & 0.443  & 0.315  & 0.404  & 0.345  & 0.394  & 0.345  & 0.394  \\
    \midrule
    \multirow{5}[4]{*}{\begin{sideways}ILI\end{sideways}} & 24 & \textcolor[rgb]{ 1,  0,  0}{\textbf{1.263 }} & \textcolor[rgb]{ 0,  0,  1}{\underline{0.742} } & 1.469  & 0.800  & 1.347  & \textcolor[rgb]{ 1,  0,  0}{\textbf{0.717 }} & \textcolor[rgb]{ 0,  0,  1}{\underline{1.319} } & 0.754  & 3.040  & 1.186  & 1.472  & 0.798  & 2.215  & 1.081  & 2.317  & 0.934  & 2.684  & 1.112  & 4.337  & 1.507  & 4.445  & 1.536  \\
       & 36 & \textcolor[rgb]{ 1,  0,  0}{\textbf{1.203 }} & \textcolor[rgb]{ 1,  0,  0}{\textbf{0.703 }} & 1.450  & 0.845  & \textcolor[rgb]{ 0,  0,  1}{\underline{1.250} } & 0.778  & 1.430  & 0.834  & 3.356  & 1.230  & 1.435  & \textcolor[rgb]{ 0,  0,  1}{\underline{0.745} } & 1.963  & 0.963  & 1.972  & 0.920  & 2.507  & 1.013  & 4.205  & 1.481  & 4.409  & 1.519  \\
       & 48 & \textcolor[rgb]{ 1,  0,  0}{\textbf{1.227 }} & \textcolor[rgb]{ 1,  0,  0}{\textbf{0.763 }} & 1.572  & 0.875  & \textcolor[rgb]{ 0,  0,  1}{\underline{1.388} } & \textcolor[rgb]{ 0,  0,  1}{\underline{0.781} } & 1.553  & 0.815  & 3.441  & 1.223  & 1.474  & 0.822  & 2.130  & 1.024  & 2.238  & 0.940  & 2.423  & 1.012  & 4.257  & 1.484  & 4.388  & 1.507  \\
       & 60 & \textcolor[rgb]{ 0,  0,  1}{\underline{1.490} } & \textcolor[rgb]{ 0,  0,  1}{\underline{0.831} } & 1.741  & 0.920  & 1.774  & 0.868  & \textcolor[rgb]{ 1,  0,  0}{\textbf{1.470 }} & \textcolor[rgb]{ 1,  0,  0}{\textbf{0.788 }} & 3.608  & 1.302  & 1.839  & 0.912  & 2.368  & 1.096  & 2.027  & 0.928  & 2.653  & 1.085  & 4.278  & 1.487  & 4.306  & 1.502  \\
\cmidrule{2-24}       & Avg & \textcolor[rgb]{ 1,  0,  0}{\textbf{1.296 }} & \textcolor[rgb]{ 1,  0,  0}{\textbf{0.760 }} & 1.558  & 0.860  & \textcolor[rgb]{ 0,  0,  1}{\underline{1.440} } & \textcolor[rgb]{ 0,  0,  1}{\underline{0.786} } & 1.443  & 0.798  & 3.361  & 1.235  & 1.555  & 0.819  & 2.169  & 1.041  & 2.139  & 0.931  & 2.567  & 1.055  & 4.269  & 1.490  & 4.387  & 1.516  \\
    \midrule
    \multicolumn{2}{c}{1st Count} & \multicolumn{2}{c}{\textcolor[rgb]{ 1,  0,  0}{\textbf{52}}} & \multicolumn{2}{c}{1} & \multicolumn{2}{c}{\textcolor[rgb]{ 0,  0,  1}{\underline{9}}} & \multicolumn{2}{c}{\textcolor[rgb]{ 0,  0,  1}{\underline{9}}} & \multicolumn{2}{c}{0} & \multicolumn{2}{c}{0} & \multicolumn{2}{c}{1} & \multicolumn{2}{c}{0} & \multicolumn{2}{c}{1} & \multicolumn{2}{c}{7} & \multicolumn{2}{c}{2} \\
    \bottomrule
    \end{tabular}%
    \label{table: Main results}
\end{table*}%

In this study, we evaluated EffiCANet against a comprehensive set of baseline models across nine datasets, using prediction lengths of 24, 36, 48, 60 for the ILI dataset, and 96, 192, 336, 720 for the others. Each model was evaluated under optimized conditions, ensuring fair and accurate comparisons across different methodologies in time series forecasting.

EffiCANet consistently delivered strong results, as summarized in \autoref{table: Main results}. Across 72 experimental evaluations, it ranked first in 51 cases and second in 13, demonstrating superior performance in most scenarios. For instance, on the ETTh2 dataset, EffiCANet achieved a 4.7\% reduction in MSE and a 2.53\% reduction in MAE compared to the second-best model. Similarly, on the ILI dataset, it achieved a 10.02\% reduction in MSE and a 3.34\% reduction in MAE. These consistent gains underscore the model's robustness and adaptability.

Analyzing the baseline models, convolution-based approaches like ModernTCN excel on datasets with shorter sequences or periodic patterns, such as Exchange and ILI, where local temporal dependencies dominate. However, they struggle with longer sequences due to limited capacity in capturing extended dependencies. Transformer-based models like PatchTST perform well on datasets with more variables or longer sequences, such as Traffic and Electricity, owing to their capability to model intricate interactions and long-term dependencies, but may underperform with irregular or shorter time series. MLP-based models, like RLinear, show efficiency on simpler datasets with fewer variables, such as the ETT series. However, they tend to fall short when dealing with complex temporal patterns and inter-variable relationships, particularly in datasets with a high number of variables, such as Traffic.

EffiCANet outperforms these models by effectively capturing both long-range dependencies and short-term dynamics through large-kernel decomposed convolutions. The inter-variable group convolution further optimizes variable interactions over close time intervals, while global temporal-variable attention dynamically focuses on relevant features. This holistic design allows EffiCANet to excel in scenarios requiring nuanced analysis of both temporal and cross-variable relationships.
\vspace{-5pt}
\subsection{Ablation Study}

\begin{table}[htbp]
  \centering
  \caption{Ablation study of removing temporal and variable-specific modules. The best results are shown in \textbf{bold}.}
  \footnotesize
  \setlength{\tabcolsep}{1.6pt}
    \begin{tabular}{ccccccccccc}
    \toprule
    Dataset & \multicolumn{2}{c}{ETTm2} & \multicolumn{2}{c}{ETTh2} & \multicolumn{2}{c}{Weather} & \multicolumn{2}{c}{Traffic} & \multicolumn{2}{c}{ILI} \\
    Metric & MSE & MAE & MSE & MAE & MSE & MAE & MSE & MAE & MSE & MAE \\
    \midrule
    Ours & \textbf{0.248 } & \textbf{0.312 } & \textbf{0.304 } & \textbf{0.366 } & \textbf{0.219 } & \textbf{0.262 } & \textbf{0.387 } & \textbf{0.278 } & \textbf{1.296 } & \textbf{0.760 } \\
    w/o TD & 0.251  & 0.315  & 0.307  & 0.371  & 0.225  & 0.268  & 0.417  & 0.297  & 1.664  & 0.857  \\
    w/o VD & 0.249  & 0.314  & 0.307  & 0.368  & 0.223  & 0.263  & 0.407  & 0.283  & 1.600  & 0.807  \\
    w TLDC & 0.250  & 0.316  & 0.308  & 0.369  & 0.228  & 0.263  & 0.409 & 0.285 & 1.671  & 0.820  \\
    w IVGC & 0.256  & 0.318  & 0.317  & 0.372  & 0.230  & 0.274  & 0.418  & 0.299  & 2.202  & 0.977  \\
    \bottomrule
    \end{tabular}%
  \label{tab:Ablation1}%
\end{table}%

\begin{table}[htbp]
  \centering
  \caption{Ablation study comparing the performance and computational efficiency between our TLDC and the standard TLC. \colorbox[rgb]{ .89,  .949,  .851}{Green} highlights denote instances where TLDC outperforms TLC, while \colorbox[rgb]{ .851,  .882,  .957}{blue} highlights indicate comparable performance.}
  \small
  \setlength{\tabcolsep}{2.2pt}
    \begin{tabular}{c|cccc|cccc}
    \toprule
    Method & \multicolumn{4}{c|}{TLDC} & \multicolumn{4}{c}{TLC} \\
    \midrule
    Metric & MSE & MAE & FLOPs & params & MSE & MAE & FLOPs & params \\
    \midrule
    ETTh1 & 0.387  & 0.414  & 0.4666  & 0.5250  & \cellcolor[rgb]{ .89,  .949,  .851}0.388  & \cellcolor[rgb]{ .851,  .882,  .957}0.414  & \cellcolor[rgb]{ .89,  .949,  .851}0.6322  & \cellcolor[rgb]{ .89,  .949,  .851}0.5447  \\
    ECL & 0.156  & 0.252  & 0.6114  & 1.8452  & \cellcolor[rgb]{ .851,  .882,  .957}0.156  & \cellcolor[rgb]{ .851,  .882,  .957}0.252  & \cellcolor[rgb]{ .89,  .949,  .851}0.8283  & \cellcolor[rgb]{ .89,  .949,  .851}2.7491  \\
    Weather & 0.219  & 0.262  & 0.0213  & 0.8114  & \cellcolor[rgb]{ .851,  .882,  .957}0.219  & 0.260  & \cellcolor[rgb]{ .89,  .949,  .851}0.0289  & \cellcolor[rgb]{ .89,  .949,  .851}0.8706  \\
    Exchange & 0.302  & 0.366  & 0.0240  & 0.0341  & \cellcolor[rgb]{ .851,  .882,  .957}0.302  & \cellcolor[rgb]{ .851,  .882,  .957}0.366  & \cellcolor[rgb]{ .89,  .949,  .851}0.0330  & \cellcolor[rgb]{ .89,  .949,  .851}0.0567  \\
    \bottomrule
    \end{tabular}%
  \label{tab:Ablation2}%
\end{table}%

To evaluate the contributions of key components in our model, we conducted an ablation study, as shown in \autoref{tab:Ablation1} and \autoref{tab:Ablation2}, across four datasets: ETTh1, Electricity, Weather, and Exchange. The analysis covers two aspects: the effect of removing temporal and variable-specific modules and a comparison between large-kernel decomposed convolution (TLDC) and standard large-kernel convolution (LKC). Results are averaged across four prediction horizons for a comprehensive assessment.

In \autoref{tab:Ablation1}, we analyze the impact of removing key modules. Replacing the temporal-specific components, including the TLDC and global temporal attention, with standard convolution (w/o TD) results in a noticeable performance drop across all datasets, underscoring the importance of capturing temporal dependencies. Similarly, when the variable-specific components, including the IVGC and global variable attention, are replaced by standard convolution (w/o VD), the model's performance degrades, particularly on datasets with numerous correlated variables like Traffic, highlighting the need for cross-variable interaction modeling.

We then test the effect of using only the TLDC or IVGC modules. While the TLDC alone maintains some stability in temporally dependent datasets, it cannot fully compensate for the lack of variable interactions. Conversely, using only the IVGC leads to a sharper performance decline, demonstrating the interdependence of temporal and variable modeling, especially in complex datasets. This also emphasizes the role of global attention, which enhances both temporal and variable convolution by refining their dependencies.

In \autoref{tab:Ablation2}, we compare the TLDC with the standard LKC. For this comparison, we calculated FLOPs (in Gigaflops) and parameter counts (in Millions) using batch sizes of 100 for the ETTh1 and Exchange datasets, and 1 for the ECL and Weather datasets. The prediction accuracy between the two methods remains nearly identical, with deviations in MAE and MSE within 0.8\%. However, the computational advantages of TLDC are substantial. For instance, on the ECL dataset, TLDC reduces parameters by 32.9\% and FLOPs by 26.2\%. This highlights the efficiency of TLDC, especially in resource-constrained environments where computational costs are crucial. The slight accuracy trade-off is offset by significant improvements in efficiency, making TLDC the superior choice for balancing performance and deployability.
\vspace{-5pt}
\subsection{Model Efficiency}

\begin{figure*}[htbp]
    \centering
        \begin{subfigure}{0.33\textwidth}
        \centering
        \includegraphics[width=\textwidth]{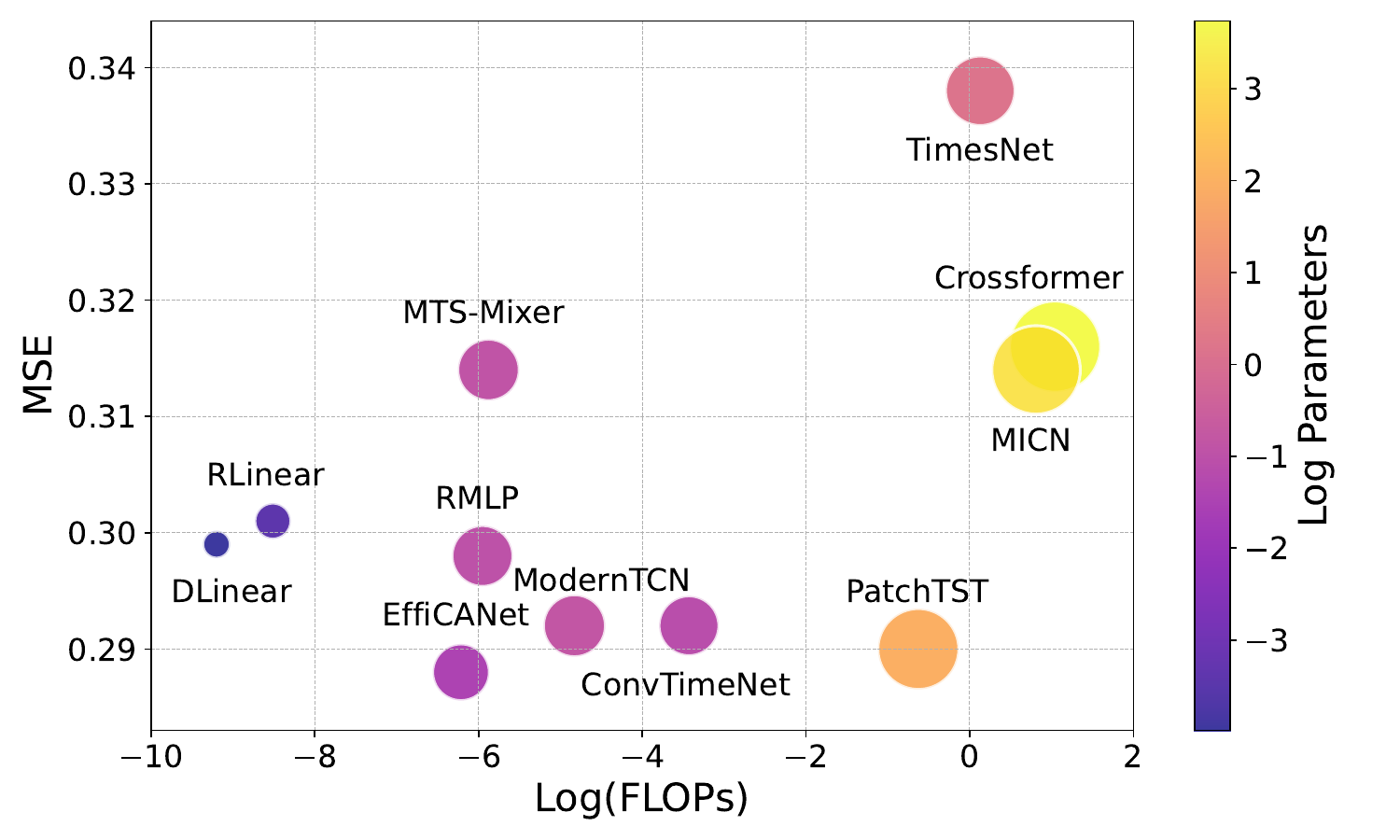}
        \caption{ETTm1 dataset}
        \label{fig:effi(a)}
    \end{subfigure}
    \begin{subfigure}{0.33\textwidth}
        \centering
        \includegraphics[width=\textwidth]{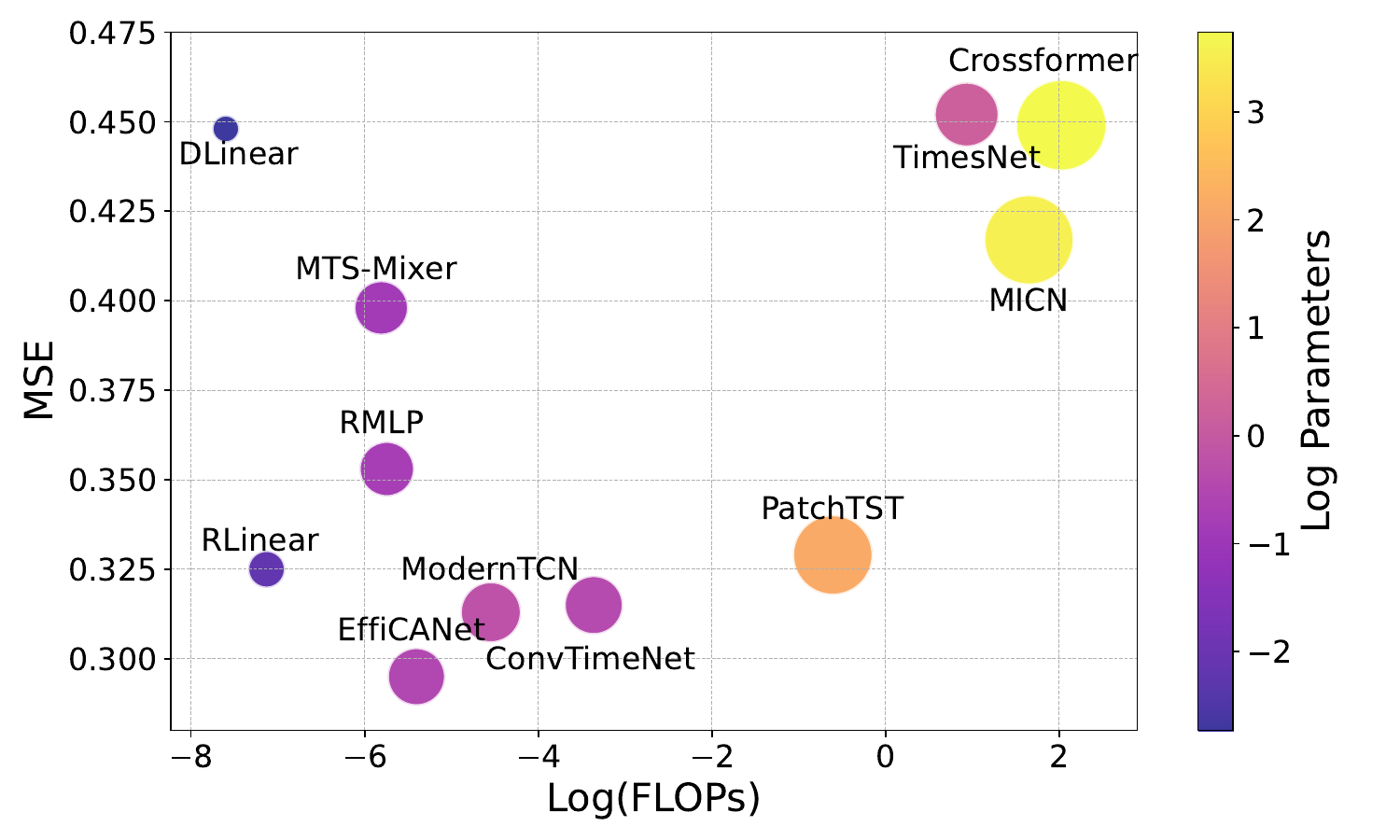}
        \caption{ETTh2 dataset}
        \label{fig:effi(b)}
    \end{subfigure}
    \begin{subfigure}{0.33\textwidth}
        \centering
        \includegraphics[width=\textwidth]{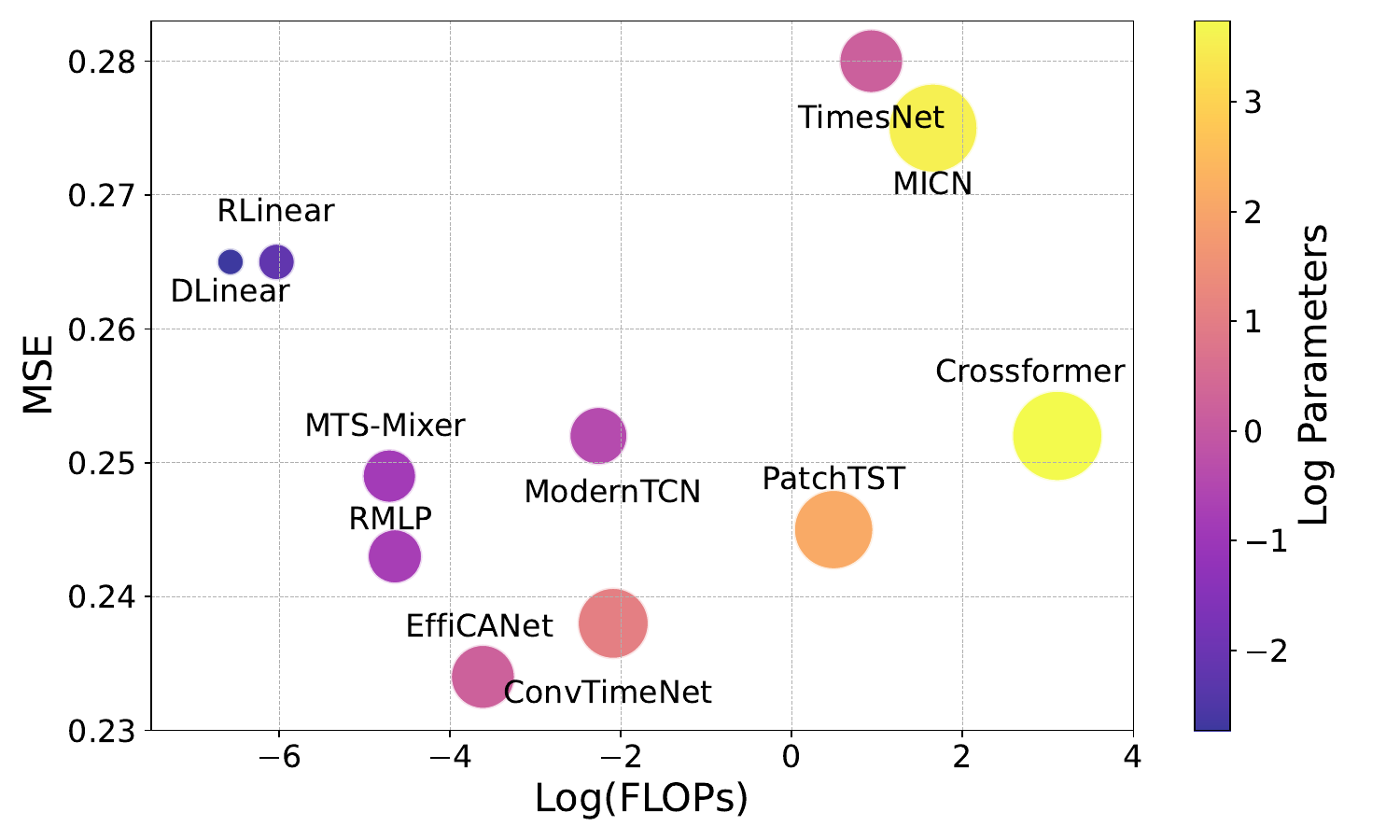}
        \caption{Weather dataset}
        \label{fig:effi(b)}
    \end{subfigure}
    \caption{Comparison of model efficiency on the ETTm1 dataset. The x-axis represents log-transformed FLOPs, while the y-axis shows MSE. Bubble size and color indicate the log-transformed number of parameters.}
    \label{fig:effi}
\end{figure*}

We evaluate the computational efficiency and prediction performance of EffiCANet against various baselines on three datasets: ETTm1 with a prediction horizon of 96, and ETTh2 and Weather with horizons of 336. The evaluation considers three key metrics: FLOPs, parameter count, and MSE. As shown in ~\autoref{fig:effi}, the results are presented using logarithmic scales for FLOPs and parameters to clearly visualize computational differences. 

MLP-based models, including DLinear, RLinear, RMLP, and MTS-Mixer, exhibit extremely low FLOPs and parameter counts. While DLinear and RLinear achieve relatively competitive MSE on simpler datasets like ETTm1, their performance declines significantly on more complex datasets with higher variable counts, such as the Weather dataset. Other MLP-based models similarly underperform compared to more advanced architectures, underscoring their limitations in capturing complex inter-variable dependencies, which is crucial for multivariate time series forecasting.

Transformer-based models, such as PatchTST and Crossformer, show a substantial increase in both FLOPs and parameters, reflecting their more complex architectures. However, this complexity does not always lead to better performance. For instance, Crossformer has high computational demands but produces relatively less accurate predictions, suggesting that its architecture may be over-parameterized for the given task. PatchTST demonstrates a better balance between complexity and performance but still suffers from high computational costs.

Convolution-based models, such as ModernTCN, ConvTimeNet, and EffiCANet, achieve a more favorable balance between efficiency and accuracy. Notably, EffiCANet achieves the lowest MSE among all models while maintaining a significantly lower computational footprint compared to Transformer models. This efficiency makes our approach particularly advantageous for real-time applications or scenarios with limited computational resources. Other convolutional models, like ModernTCN and ConvTimeNet, also show strong performance with relatively low FLOPs, further underscoring the suitability of convolutional architectures for multivariate forecasting tasks.
\vspace{-5pt}
\subsection{Parameter Sensitivity}

\begin{figure*}[htbp]
    \centering
        \begin{subfigure}{0.24\textwidth}
        \centering
        \includegraphics[width=\textwidth]{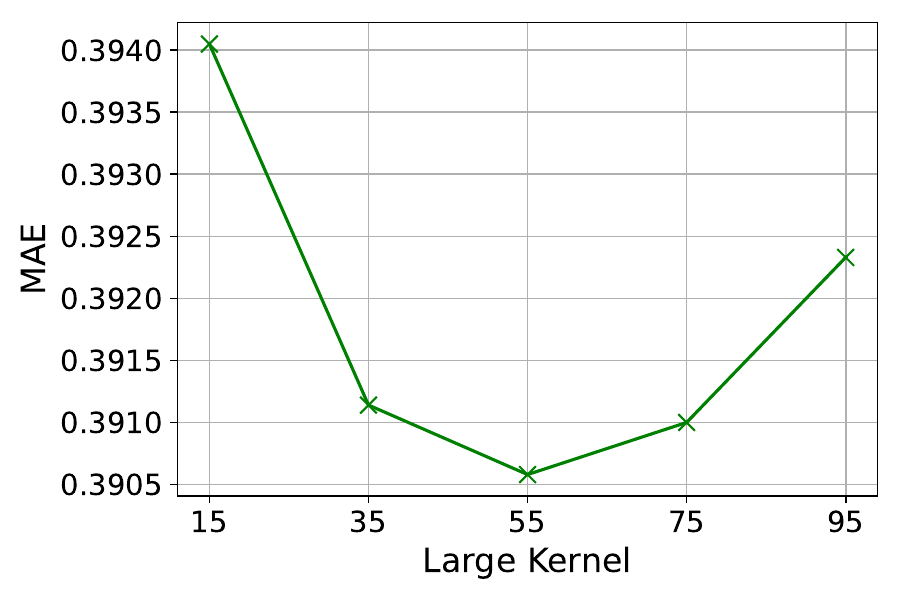}
        \caption{MAE vs Large Kernel}
        \label{fig:para(a)}
    \end{subfigure}
    \begin{subfigure}{0.24\textwidth}
        \centering
        \includegraphics[width=\textwidth]{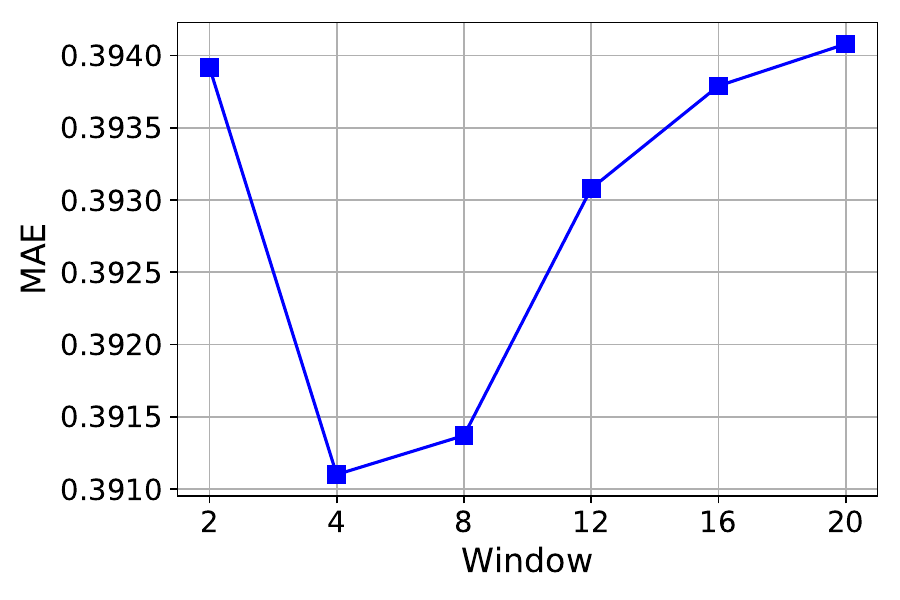}
        \caption{MAE vs Window}
        \label{fig:para(b)}
    \end{subfigure}
    \begin{subfigure}{0.24\textwidth}
        \centering
        \includegraphics[width=\textwidth]{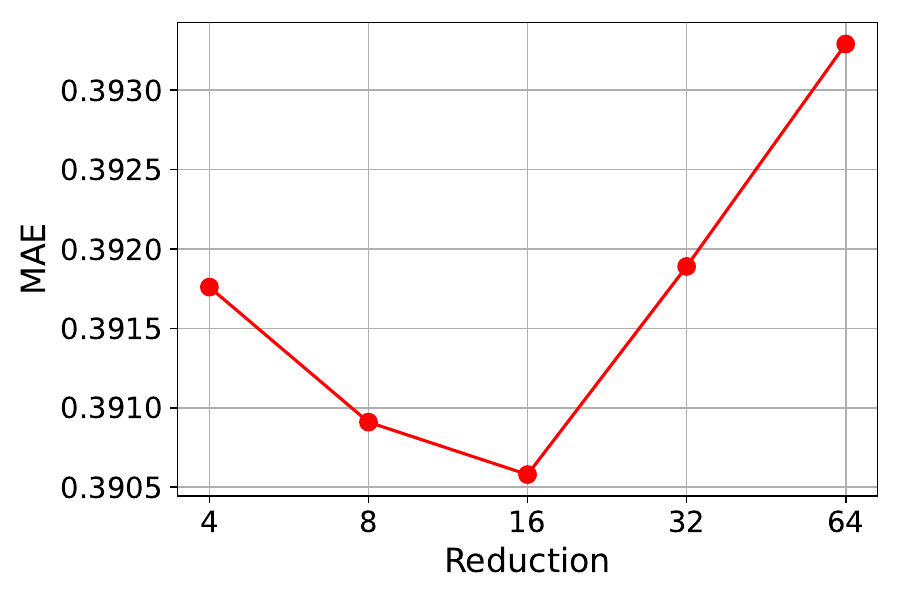}
        \caption{MAE vs Reduction}
        \label{fig:para(c)}
    \end{subfigure}
    \begin{subfigure}{0.24\textwidth}
        \centering
        \includegraphics[width=\textwidth]{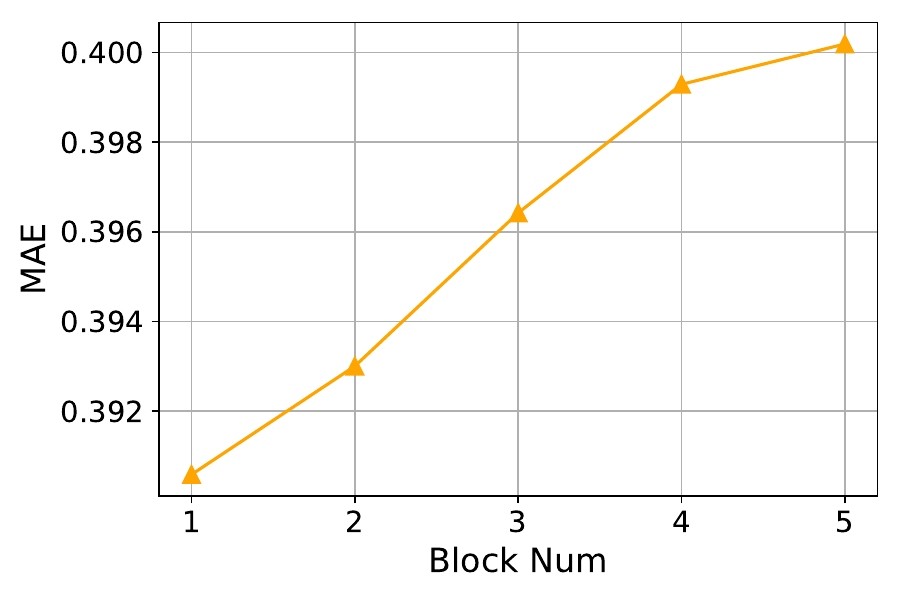}
        \caption{MAE vs Block Numbers}
        \label{fig:para(d)}
    \end{subfigure}
    \caption{Parameter Sensitivity.}
    \label{fig:para}
\end{figure*}

We perform parameter sensitivity analysis on the ETTh1 dataset with a prediction horizon of 96, using MAE as the evaluation metric. The focus is on four key parameters: the large-kernel size $K$  and its corresponding dilation rate $d$ in the TLDC module, the time window size $W$ in the IVGC module, the reduction ratio $r$ in the GTVA module, and the number of blocks $L$ in the model's architecture.

For the large-kernel size, as shown in ~\autoref{fig:para(a)}, we vary the $K$ from 15 to 95, with $d$ set accordingly to 1, 3, 5, 7, and 9. The MAE decreases as $K$ increases, reaching its minimum at $K = 55$. Beyond this, performance declines, likely due to over-smoothing of the temporal features. ~\autoref{fig:para(b)} explores the time window size, revealing that $W = 4$ is optimal. Smaller windows may miss temporal dependencies, while larger windows appear to introduce noise or redundancy, resulting in diminished performance.

In ~\autoref{fig:para(c)}, we find that a moderate $r=16$ provides the best balance, suggesting it effectively reduces dimensionality without losing key information. Both lower and higher ratios degrade performance, either due to retaining too much irrelevant information or over-compressing the feature space. ~\autoref{fig:para(d)} shows that using more than one block leads to overfitting, with minimal accuracy gains. A single block captures core dependencies, and limiting block count ensures model simplicity and better generalization. While these results provide optimal settings for the ETTh1 dataset, different datasets may require adjustments.
\vspace{-3pt}
\subsection{Model Scalability}

\begin{figure*}[htbp]
    \centering
        \begin{subfigure}{0.24\textwidth}
        \centering
        \includegraphics[width=\textwidth]{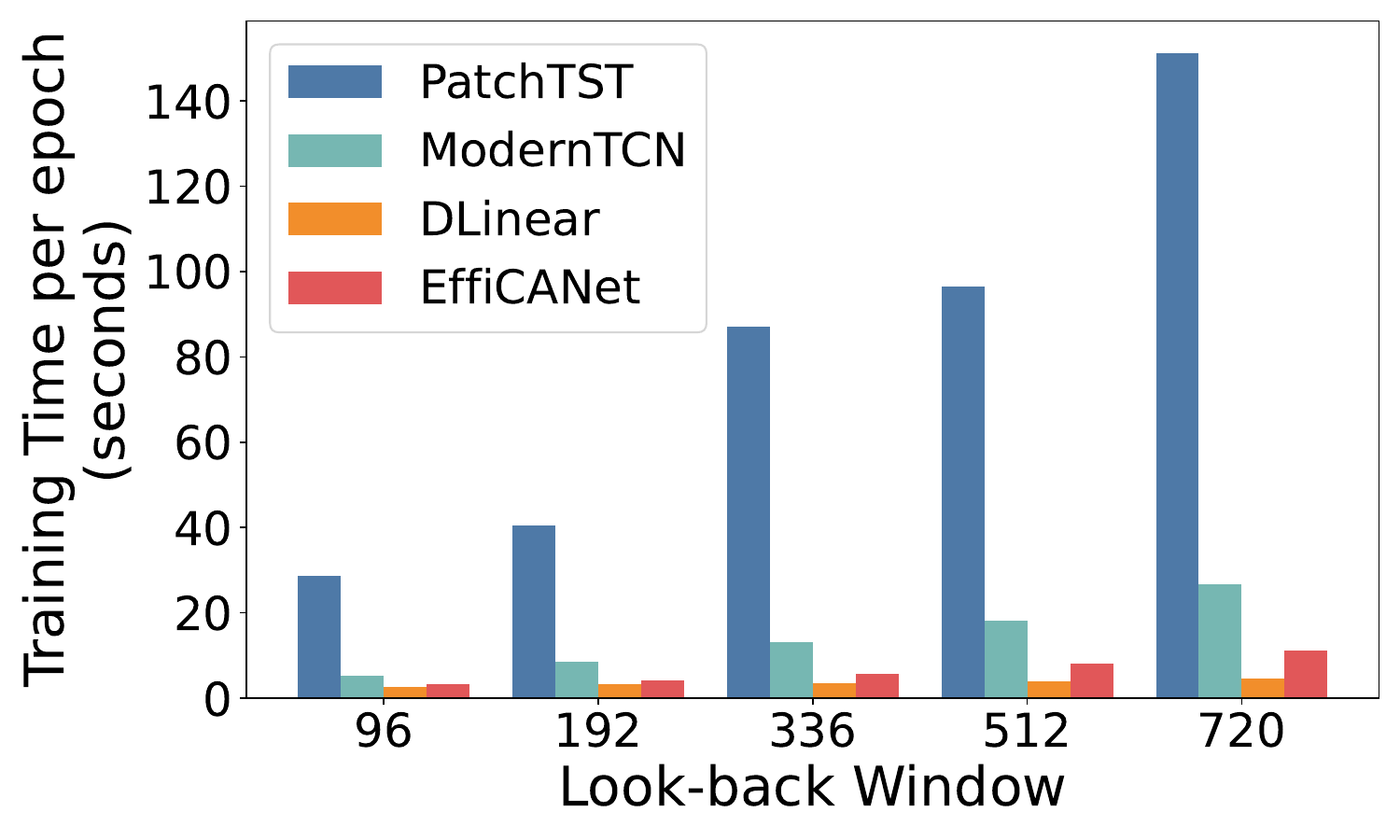}
        \caption{Training time vs $H$}
        \label{fig:scale(a)}
    \end{subfigure}
    \begin{subfigure}{0.24\textwidth}
        \centering
        \includegraphics[width=\textwidth]{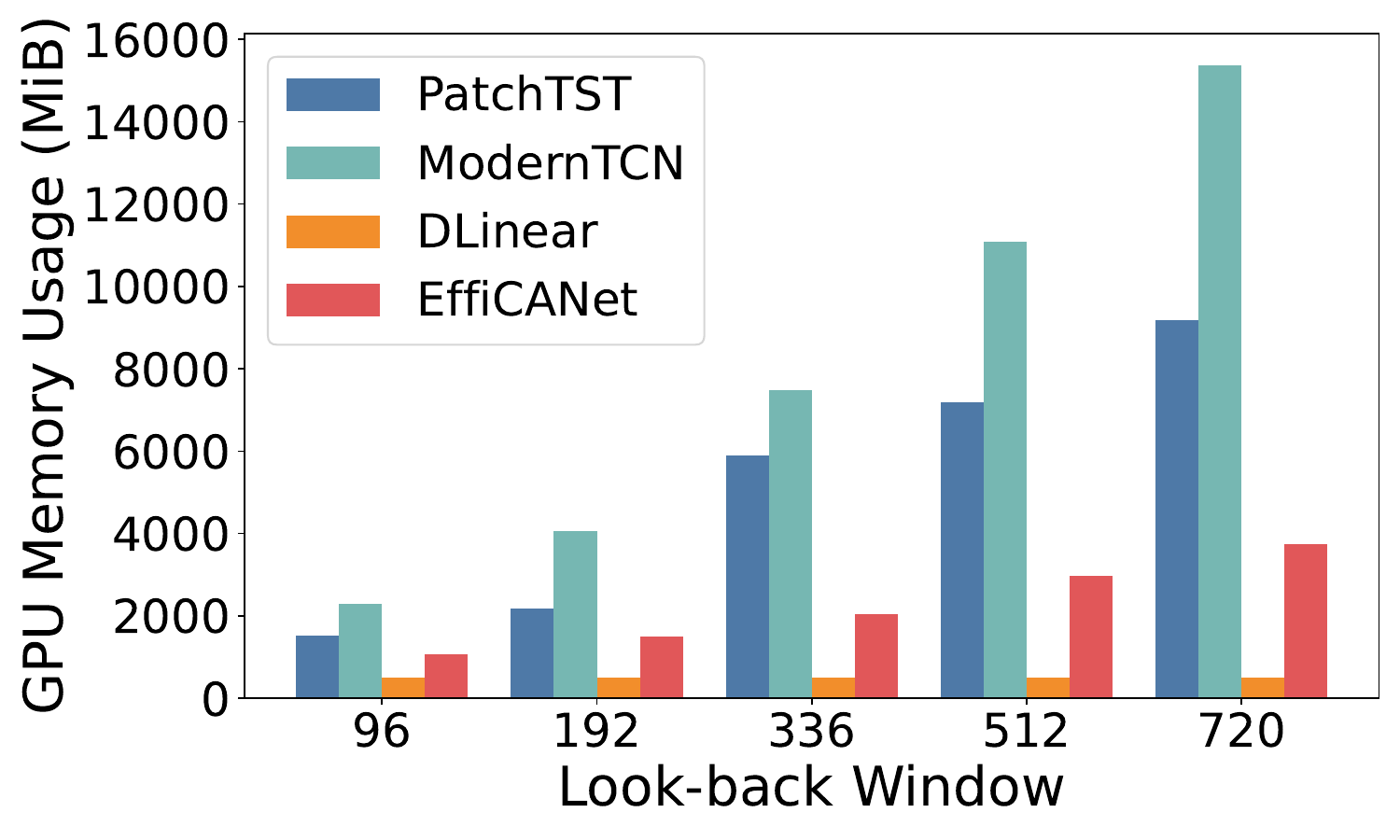}
        \caption{GPU memory vs $H$}
        \label{fig:scale(b)}
    \end{subfigure}
    \begin{subfigure}{0.24\textwidth}
        \centering
        \includegraphics[width=\textwidth]{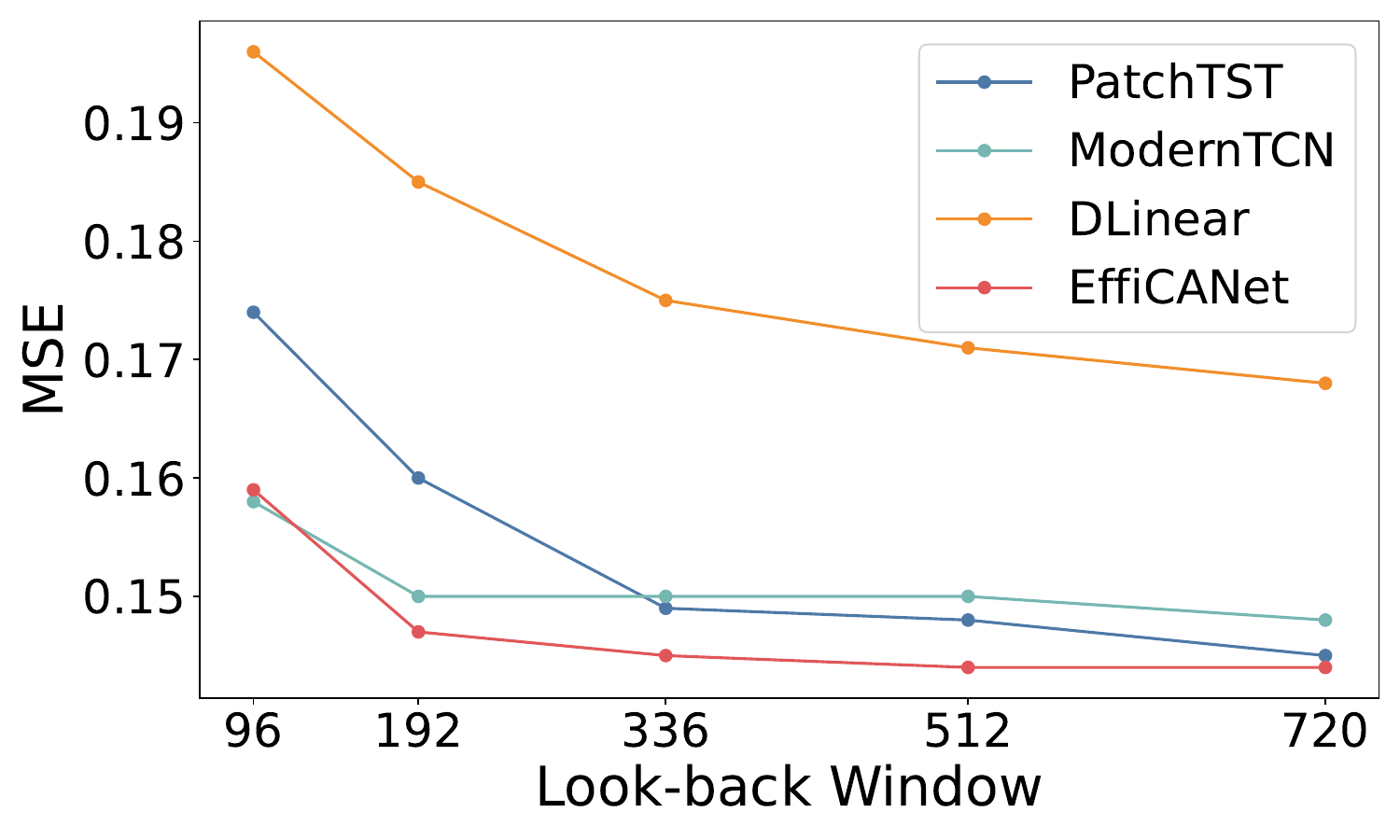}
        \caption{MSE vs $H$}
        \label{fig:scale(c)}
    \end{subfigure}
    \begin{subfigure}{0.24\textwidth}
        \centering
        \includegraphics[width=\textwidth]{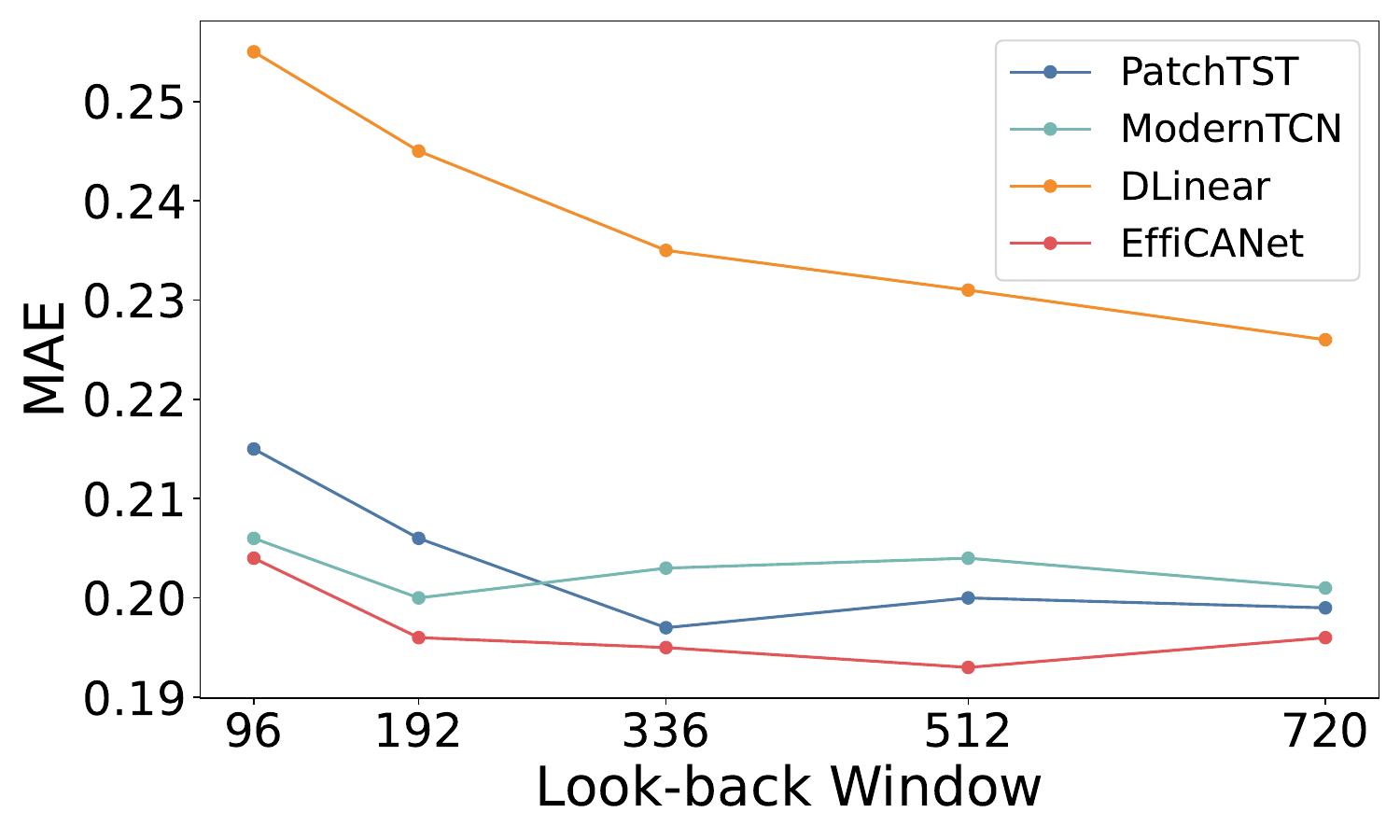}
        \caption{MAE vs $H$}
        \label{fig:scale(d)}
    \end{subfigure}
    \caption{Model Scalability across difference look-back windows.}
    \label{fig:scale}
\end{figure*}

We evaluate EffiCANet’s scalability on the Weather dataset with sequence lengths of 96, 192, 336, 512, and 720, and a fixed prediction horizon of 96. We compare EffiCANet to three baseline models — ModernTCN, PatchTST, and DLinear — representing convolutional, Transformer, and linear architectures. The comparison includes training time per epoch, GPU memory usage, MSE, and MAE.

In \autoref{fig:scale(a)} and \autoref{fig:scale(b)}, EffiCANet consistently achieves shorter training times and lower GPU memory usage than PatchTST and ModernTCN, with the gap widening as sequence length increases. DLinear, while resource-efficient, exhibits higher error rates, highlighting its limitations in handling complex temporal dependencies. \autoref{fig:scale(c)} and \autoref{fig:scale(d)} show the MSE and MAE across varying sequence lengths. EffiCANet maintains low and stable error rates as sequence length increases, outperforming the baselines in predictive accuracy. PatchTST also shows improving accuracy with longer sequences but at a significantly higher computational cost. Overall, these results demonstrate EffiCANet’s balance of computational efficiency and accuracy across different input lengths.
\vspace{-15pt}
\subsection{Visualization}

\begin{figure*}[htbp]
    \centering
        \begin{subfigure}{0.195\textwidth}
        \centering
        \includegraphics[width=\textwidth]{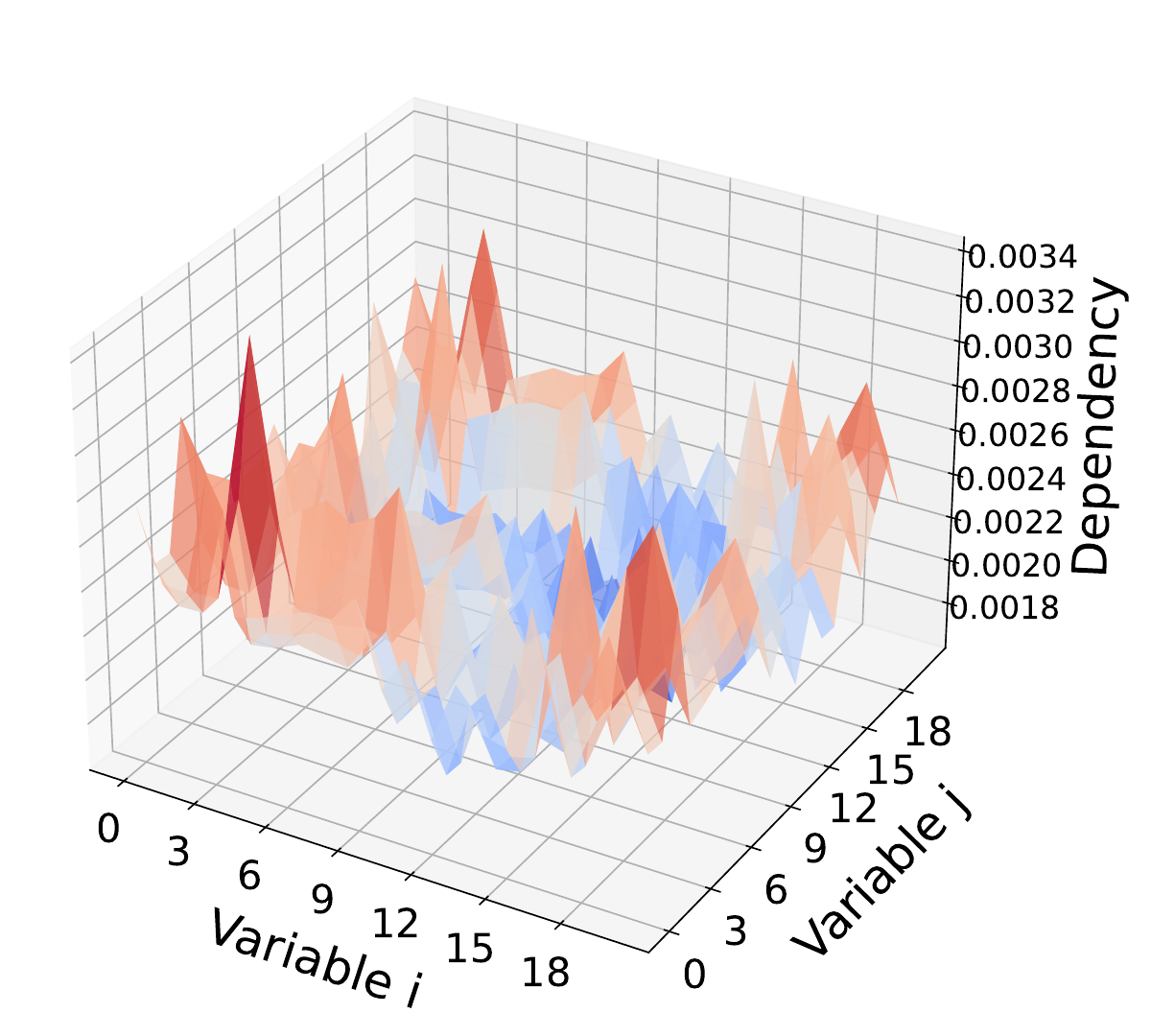}
        \caption{Time window 1}
        \label{fig:visual(a)}
    \end{subfigure}
    \begin{subfigure}{0.195\textwidth}
        \centering
        \includegraphics[width=\textwidth]{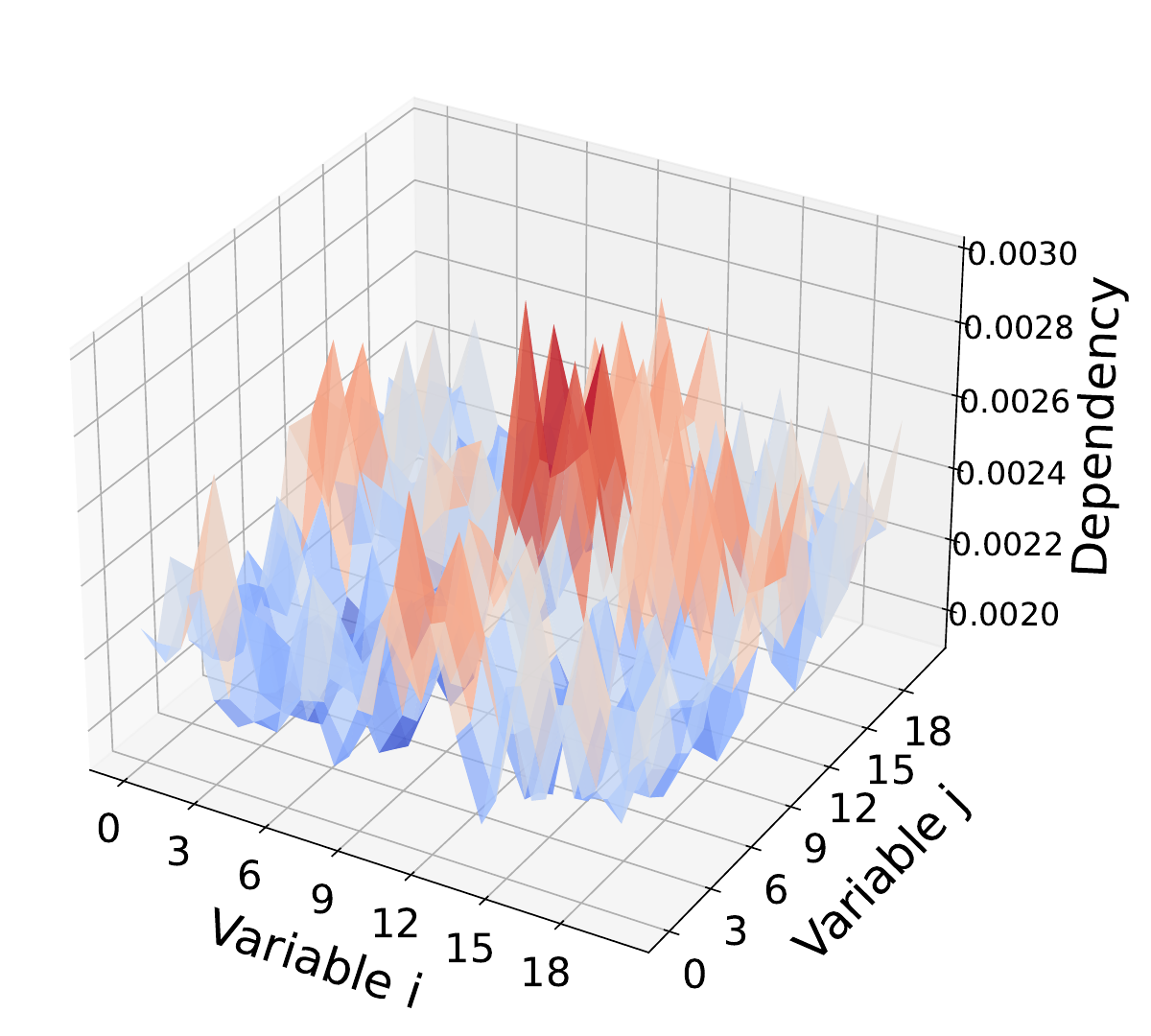}
        \caption{Time window 2}
        \label{fig:visual(b)}
    \end{subfigure}
    \begin{subfigure}{0.195\textwidth}
        \centering
        \includegraphics[width=\textwidth]{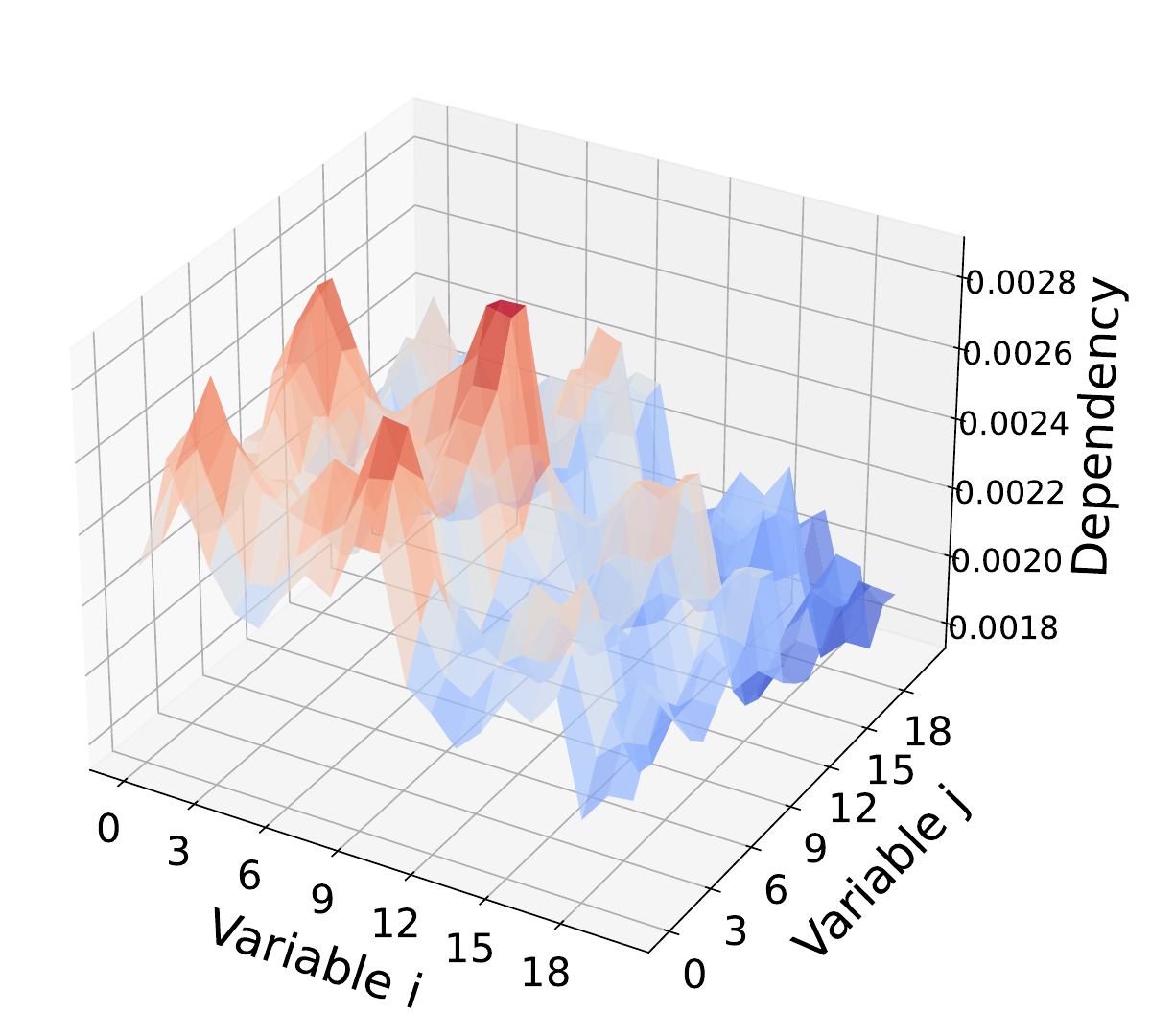}
        \caption{Time window 3}
        \label{fig:visual(c)}
    \end{subfigure}
    \begin{subfigure}{0.195\textwidth}
        \centering
        \includegraphics[width=\textwidth]{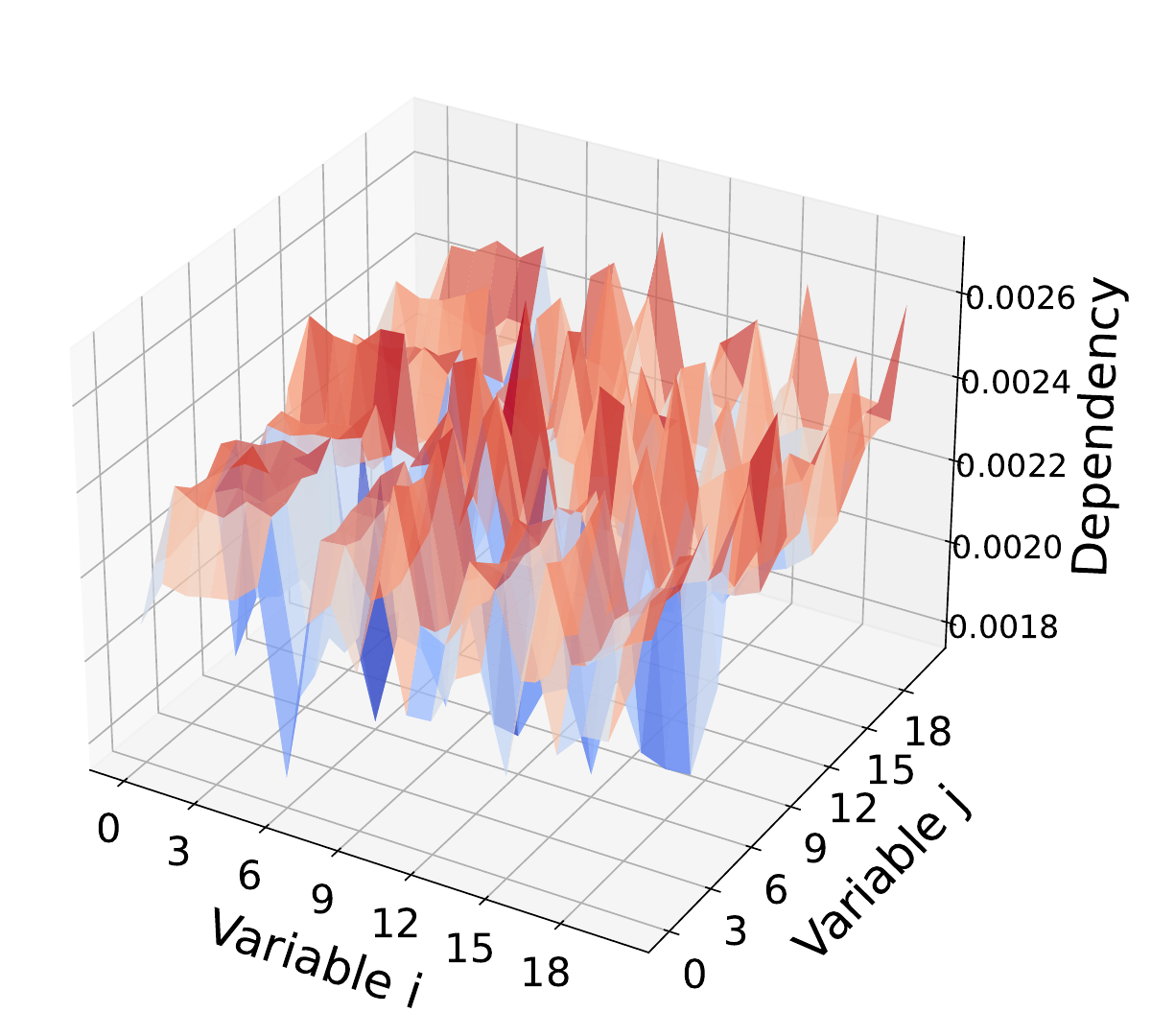}
        \caption{Time window 4}
        \label{fig:visual(d)}
    \end{subfigure}
    \begin{subfigure}{0.195\textwidth}
        \centering
        \includegraphics[width=\textwidth]{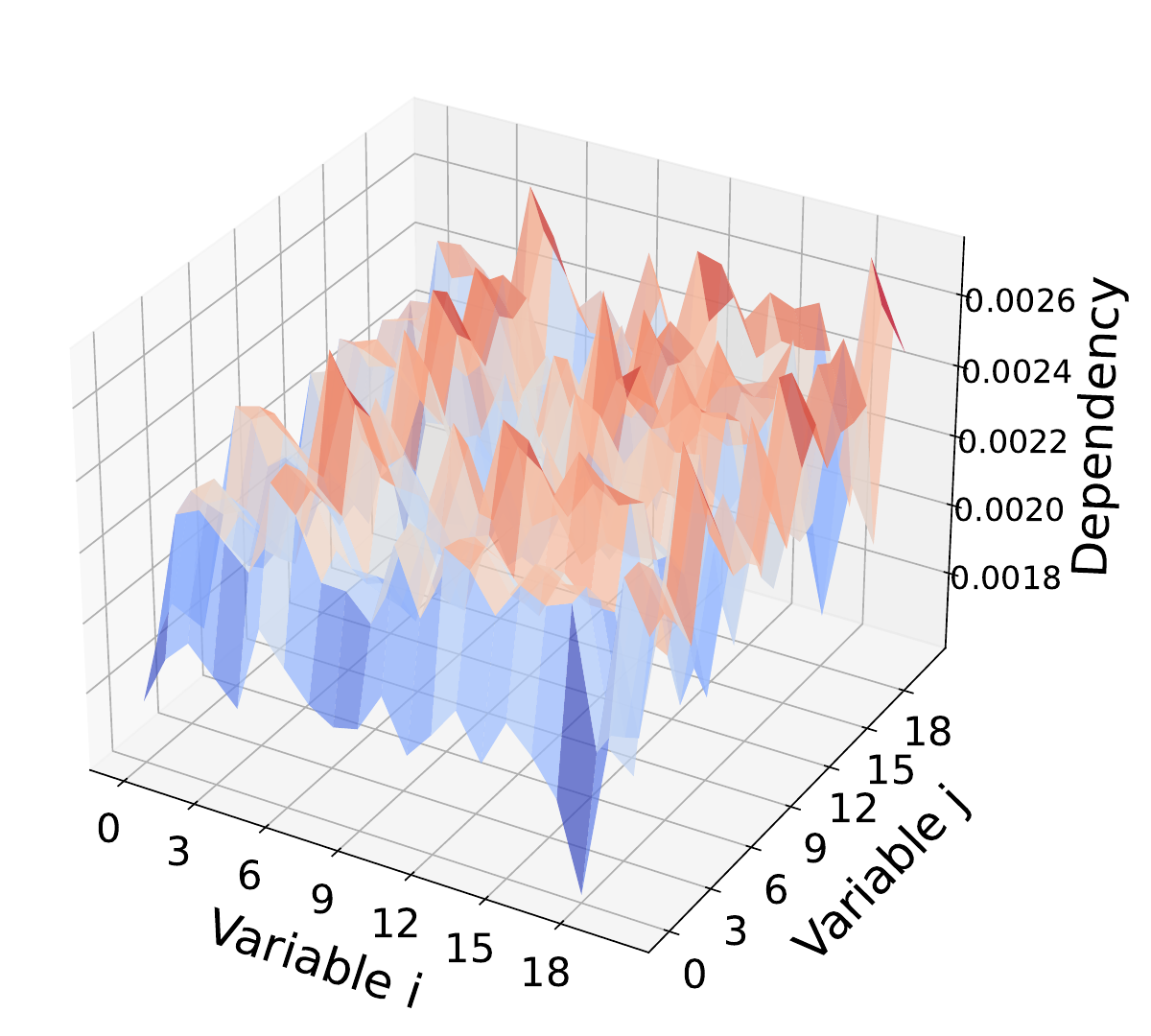}
        \caption{Time window 5}
        \label{fig:visual(e)}
    \end{subfigure}
    \caption{Variable dependency across various time windows.}
    \label{fig:visual}
\end{figure*}

\begin{figure*}[htbp]
    \centering
        \begin{subfigure}{0.24\textwidth}
        \centering
        \includegraphics[width=\textwidth]{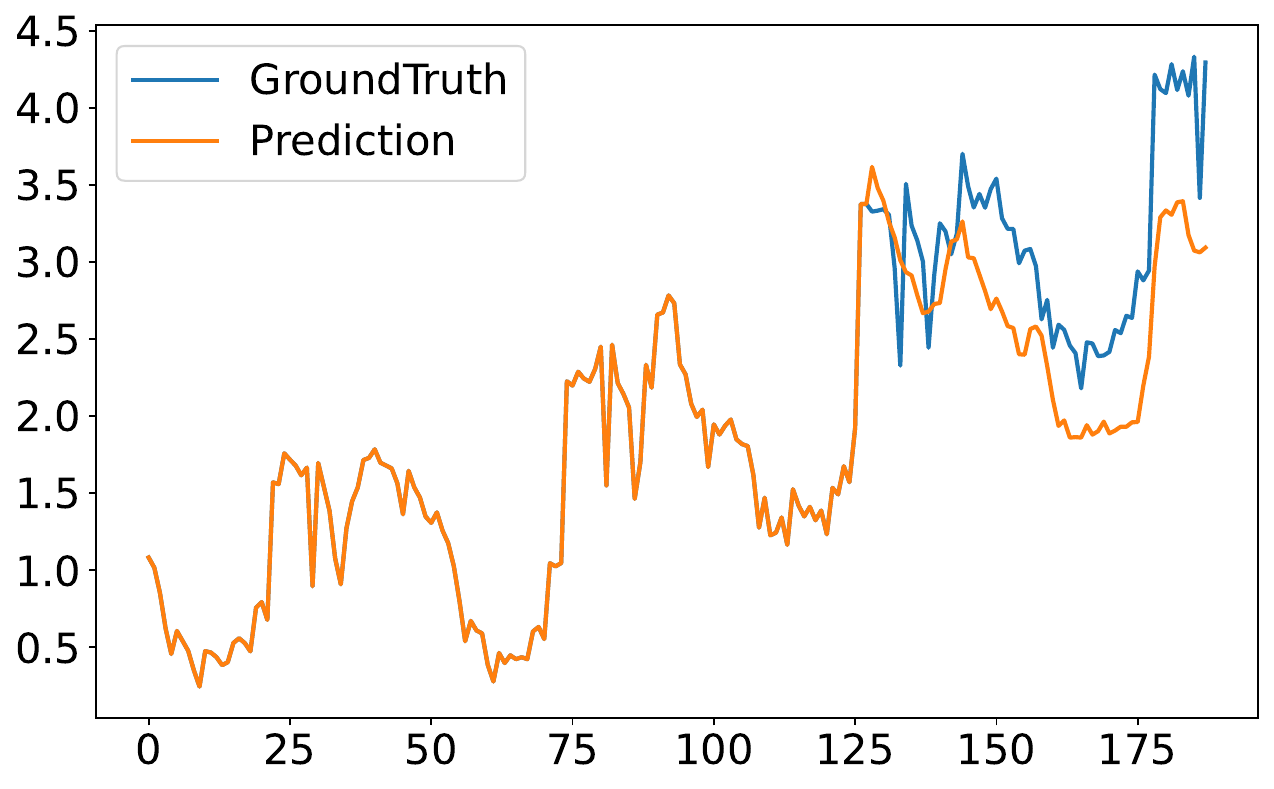}
        \caption{EffiCANet}
        \label{fig:vis(a)}
    \end{subfigure}
    \begin{subfigure}{0.24\textwidth}
        \centering
        \includegraphics[width=\textwidth]{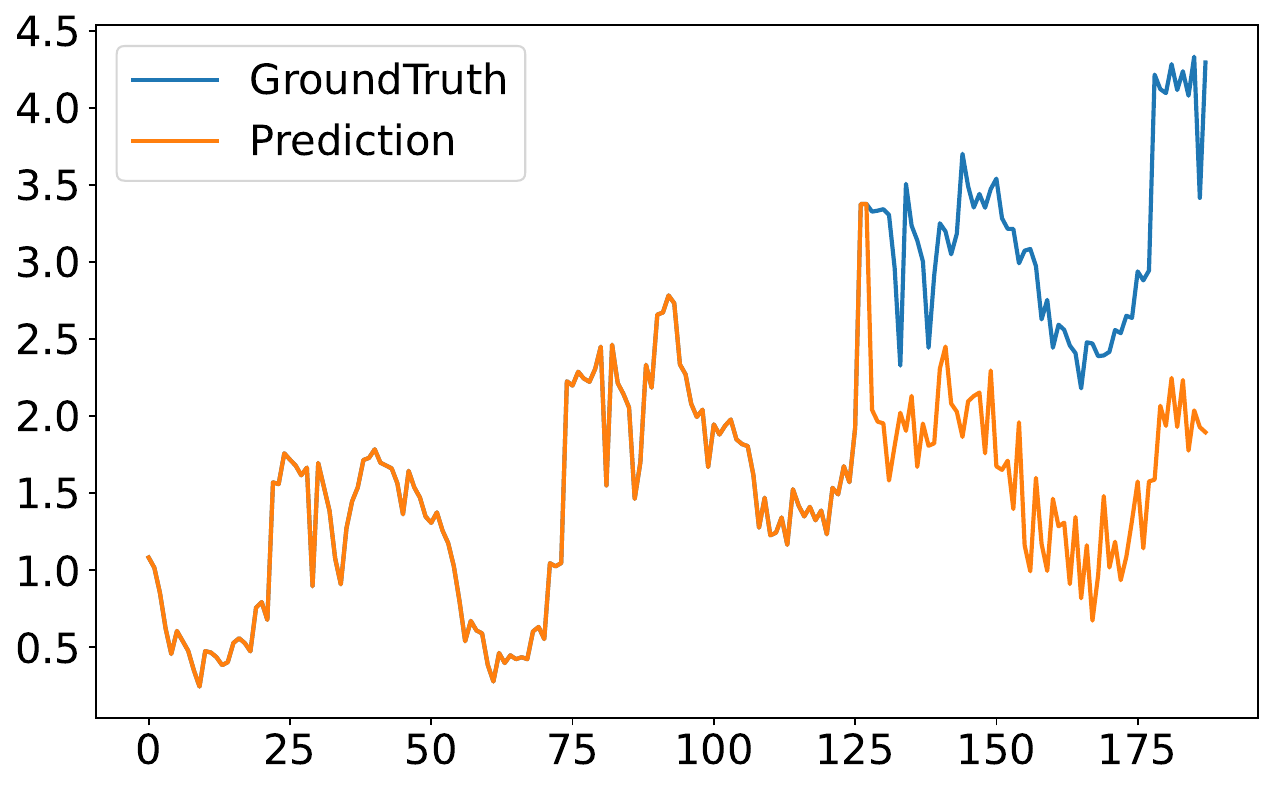}
        \caption{ModernTCN}
        \label{fig:vis(b)}
    \end{subfigure}
    \begin{subfigure}{0.24\textwidth}
        \centering
        \includegraphics[width=\textwidth]{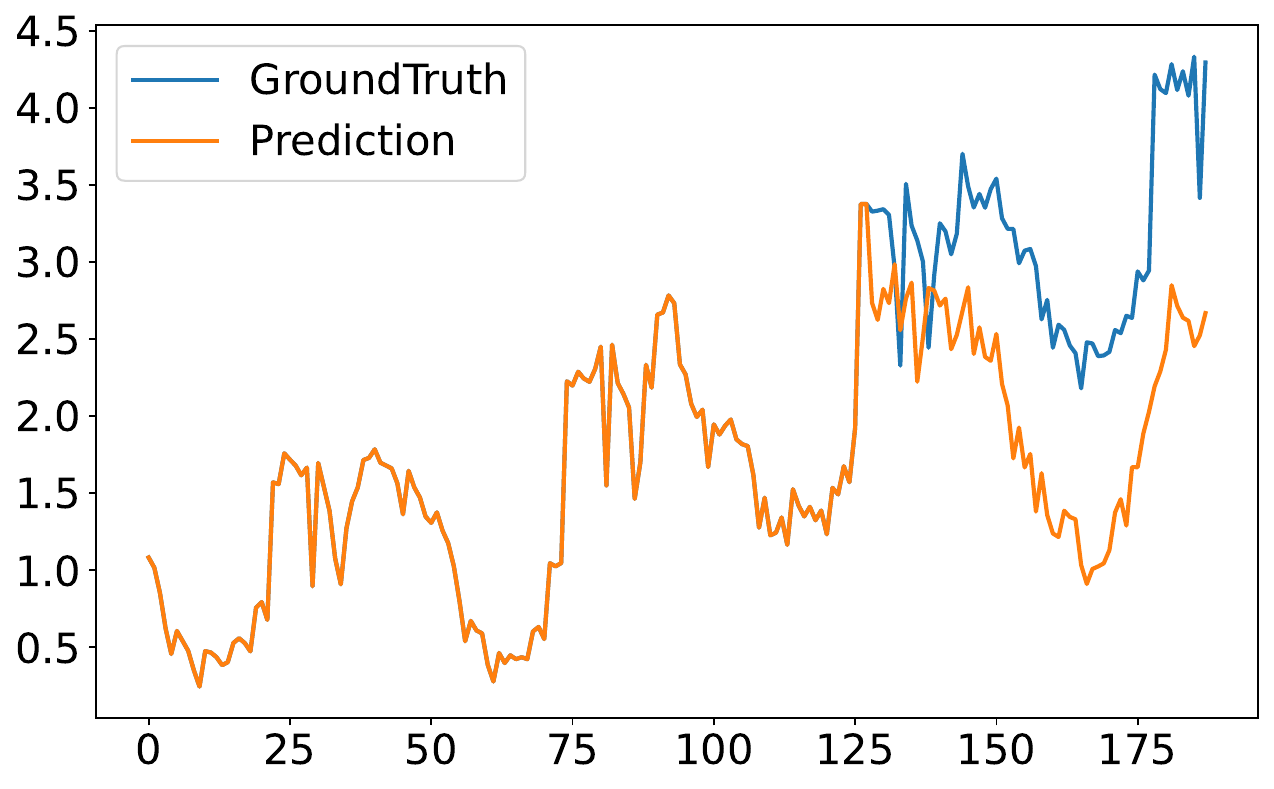}
        \caption{PatchTST}
        \label{fig:vis(c)}
    \end{subfigure}
    \begin{subfigure}{0.24\textwidth}
        \centering
        \includegraphics[width=\textwidth]{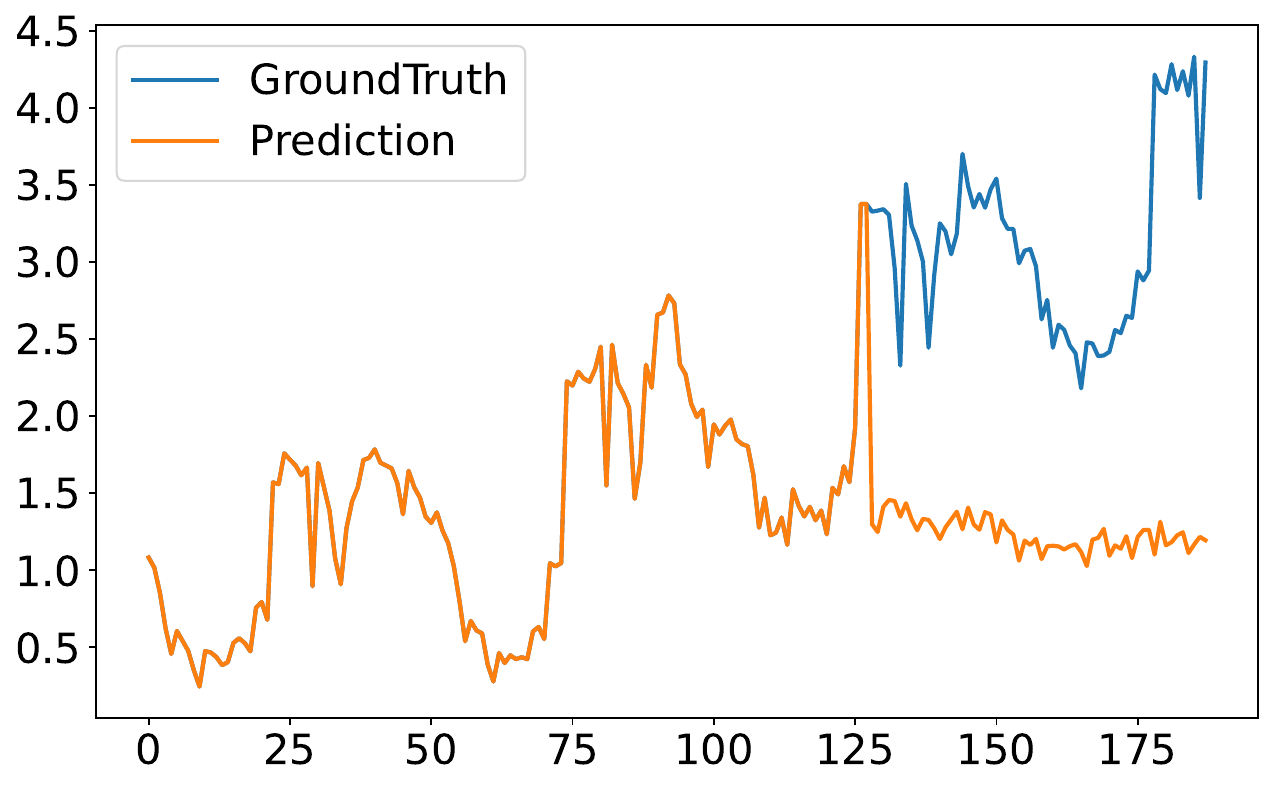}
        \caption{DLinear}
        \label{fig:vis(d)}
    \end{subfigure}
    \caption{ Forecasting visualization comparison on ILI dataset.}
    \label{fig:vis}
\end{figure*}

\subsubsection{Visualization of variable dependency}
To interpret the inter-variable dependencies captured by EffiCANet, we conducted a visualization analysis on the Weather dataset with a prediction horizon of 96. Using convolutional weights from the IVGC module across five consecutive time windows, we computed dependency values that represent relationships between variables within each window. Specifically, the product of convolutional weights for each variable pair reflects their mutual influence during the convolution operation. Summing these products provides an aggregated measure of inter-variable dependency, capturing how variables interact over time. Each subplot in \autoref{fig:visual} illustrates the dependency distribution across variables within a specific time window, providing insights into how interactions evolve dynamically.

The visualizations reveal that inter-variable dependencies fluctuate across time windows, indicating dynamic relationships that change with recent history. For example, we observe strong correlations between certain variable pairs in earlier time windows, likely reflecting short-term interactions. As time progresses, these relationships evolve, with some weakening while others emerging, revealing complex temporal patterns. This behavior mirrors real-world weather dynamics, where factors like temperature, humidity, and wind speed exhibit variable interdependence across time.

The group convolutions in the IVGC module are key to this adaptive modeling, as they localize dependency modeling within small, time-varying groups of variables. Unlike conventional approaches that assume static dependencies, our model dynamically responds to changes in variable interactions across time windows, effectively capturing both transient and persistent dependencies. This adaptability improves the model’s ability to handle time-sensitive dependencies, contributing to more accurate predictions. 

\subsubsection{Visualization of forecasting results}

We visualize the forecasting performance of EffiCANet and three baseline models — ModernTCN, PatchTST, and DLinear — on the ILI dataset with a prediction horizon of 60 timesteps. This assessment aims to assess each model’s ability to capture trends and fluctuations relative to the ground truth.

As shown in \autoref{fig:vis}, EffiCANet closely aligns with the ground truth, accurately capturing both smooth and abrupt changes in the data, especially during periods of rapid variation. While ModernTCN and PatchTST follow the general trends, they struggle to adapt to short-term fluctuations, resulting in underestimation of key peaks and troughs. DLinear exhibits the weakest performance, failing to capture key patterns and over-smoothing most of the predicted values.

\section{Conclusion}\label{Conclusion}

In this work, we propose a novel approach for multivariate time series forecasting through three specialized modules that effectively capture both temporal and inter-variable dependencies. The Temporal Large-kernel Decomposed Convolution (TLDC) module efficiently models short- and long-term temporal dependencies by decomposing large kernels, enabling scalable handling of extended sequences. The Inter-Variable Group Convolution (IVGC) module adapts to dynamic inter-variable relationships by learning localized interactions within time windows, capturing critical time-sensitive dependencies. The Global Temporal-Variable Attention (GTVA) module further enhances the model by jointly attending to temporal and variable dimensions, providing a refined contextual understanding that improves both forecasting accuracy and adaptability to complex patterns. Future work will focus on optimizing the model for non-stationary time series and enhancing interpretability to broaden its application to complex, dynamic systems.



\balance
\bibliographystyle{ACM-Reference-Format}
\bibliography{sample}

\end{sloppypar}
\end{document}